\documentclass[journal,twoside,web]{ieeecolor}
\usepackage{generic}
\usepackage{cite}
\usepackage{multirow}
\usepackage{amsmath,amssymb,amsfonts}
\usepackage{algorithmic}
\usepackage{graphicx}
\usepackage{algorithm,algorithmic}
\usepackage{hyperref}
\usepackage{gensymb}
\hypersetup{hidelinks=true}
\usepackage{textcomp}
\usepackage{booktabs}   
\usepackage{siunitx}
\usepackage{orcidlink}
\usepackage[moderate,tracking=normal]{savetrees}
\usepackage[table]{xcolor}
\usepackage{siunitx}
\def\BibTeX{{\rm B\kern-.05em{\sc i\kern-.025em b}\kern-.08em
    T\kern-.1667em\lower.7ex\hbox{E}\kern-.125emX}}
\begin{document}
\title{Towards Objective Obstetric Ultrasound Assessment: Contrastive Representation Learning for Fetal Movement Detection}
\author{%
  Talha Ilyas\orcidlink{0000-0002-4168-2998},
  Duong Nhu,
  Allison Thomas,
  Arie Levin\orcidlink{0000-0003-4352-1933},
  Lim Wei Yap,
  Shu Gong,
  David Vera Anaya,
  Yiwen Jiang,
  Deval Mehta\orcidlink{0000-0002-6907-7589},
  Ritesh Warty,
  Vinayak Smith,
  Maya Reddy,
  Euan Wallace,
  Wenlong Cheng\orcidlink{0000-0002-2346-4970}, Senior Member, IEEE,
  Zongyuan Ge\orcidlink{0000-0002-5880-8673}, Senior Member, IEEE,
  Faezeh Marzbanrad\orcidlink{0000-0003-0551-1611}, Senior Member, IEEE.
  \thanks{Manuscript received May 27, 2025; supported by NHMRC Ideas Grant (APP2004444) and Investigator Grant (APP2010154). S. Gong is supported by Jack Brockhoff Foundation Grant 4659–2019.}
    \thanks{Talha Ilyas is with the Department of Electrical and Computer Systems Engineering, Monash University, Clayton, VIC 3168, Australia. Talha Ilyas, Duong Nhu, Yiwen Jiang, Deval Mehta, Zongyuan Ge, and Faezeh Marzbanrad are with the AIM for Health Lab, Faculty of Information Technology, Monash University, Clayton, VIC 3168, Australia; Lim Wei Yap is with the School of Biomedical Engineering, University of Sydney, Darlington, NSW 2008, Australia; Wenlong Cheng is with the Department of Chemical \& Biological Engineering, Monash University, Clayton, VIC 3168, Australia; Shu Gong, David Vera Anaya, Allison Thomas, Ritesh Warty, Vinayak Smith, Maya Reddy, and Euan Wallace are with the Department of Obstetrics \& Gynaecology, Monash University, Clayton, VIC 3168, Australia; Arie Levin is with the Department of Chemical \& Biological Engineering, Monash University, Clayton, VIC 3168, Australia.}
    \thanks{E-mail:
    \{talha.ilyas, duong.nhu, yiwen.jiang, arie.levin, deval.mehta, zongyuan.ge, faezeh.marzbanrad, ritesh.warty, wenlong.cheng\}@monash.edu;
    \{limwei.yap, aswandi.wibrianto, qinhao.li, rgao0775, yan.lu3, jane.limas, sharon.mccracken, arnold.ju\}@sydney.edu.au;
    shu.gong@csu.edu.cn;
    \{davidf.veraanaya, smith.vinayak\}@gmail.com;
    alli.thomas@monashivfgroup.com.}
}
\maketitle
\begin{abstract}
Accurate fetal movement (FM) detection is essential for assessing prenatal health, as abnormal movement patterns can indicate underlying complications such as placental dysfunction or fetal distress. Traditional methods—including maternal perception and cardiotocography (CTG)—suffer from subjectivity and limited accuracy. To address these challenges, we propose Contrastive Ultrasound Video Representation Learning (CURL), a novel self-supervised learning framework for FM detection from extended fetal ultrasound video recordings. Our approach leverages a dual-contrastive loss, incorporating both spatial and temporal contrastive learning, to learn robust motion representations. Additionally, we introduce a task-specific sampling strategy, ensuring the effective separation of movement and non-movement segments during self-supervised training, while enabling flexible inference on arbitrarily long ultrasound recordings through a probabilistic fine-tuning approach. Evaluated on an in-house dataset of 92 subjects, each with 30-minute ultrasound sessions, CURL achieves a sensitivity of 78.01\% and an AUROC of 81.60\%, demonstrating its potential for reliable and objective FM analysis. These results highlight the potential of self-supervised contrastive learning for fetal movement analysis, paving the way for improved prenatal monitoring and clinical decision-making. Our code is available at: https://github.com/Mr-TalhaIlyas/CURL/.
\end{abstract}


\section{Introduction}
\label{sec:introduction}
\IEEEPARstart{E}{nsuring} maternal and fetal well-being is a cornerstone of obstetric care, with significant implications for both individual families and public health systems \cite{1}. Contemporary challenges such as declining birth rates, rising infertility, and increasing fetal mortality underscore the need for improved monitoring methods \cite{2,3}. Among the many indicators of prenatal health, fetal movement (FM) is a critical biomarker that reflects the development of the central nervous system and musculoskeletal function. Abnormal FM patterns are associated with complications including placental dysfunction, fetal distress, and intrauterine growth restriction \cite{4,5}. Despite its importance, traditional reliance on maternal perception for FM assessment is hindered by subjectivity and various confounding factors like placental position, fetal orientation, and maternal body mass index \cite{dutton2012predictors}.

Over the years, multiple sensor and signal modalities have been applied to FM assessment, including time-domain and wavelet analyses of Doppler and ultrasound signals, pressure‐sensor recordings, and heart-rate-based monitoring such as cardiotocography (CTG) \cite{wu2013research,abeywardhana2018time}. Early systems employing maternal abdominal pressure sensors or CTG—both providing one-dimensional surrogates of FM—were informative but limited by indirect measurement and the brief duration of ultrasound examinations \cite{xu2023development,zhu2016understanding}. More recent efforts using wearable accelerometers, fetal electrocardiography (fECG), and fetal magnetocardiography (fMCG)—all heart-rate–based and thus indirect—have improved detection sensitivity; however, these modalities still struggle with signal noise and interference from maternal activity \cite{nishihara2008long,mesbah2021automatic,avci2017tracking,lutter2011indices}.

The rapid evolution of artificial intelligence (AI) has paved the way for a transformative approach in fetal ultrasound imaging. Traditional ultrasound assessments, while invaluable for anatomical and functional measurements, typically focus on static parameters such as head biometry or single‐frame movement snapshots \cite{avci2017tracking}. In contrast, video analysis of ultrasound recordings enables extraction of rich spatiotemporal features from each scan, reducing operator dependency and subjectivity. This approach enhances fetal assessment by leveraging motion cues within routine ultrasound sessions,  without the need for additional processes or wearable devices  \cite{stanojevic2023fetal}.

A key strength of our approach lies in its practical applicability. With the advent of portable ultrasound devices—such as the Butterfly iQ and other handheld systems that connect to Android or iOS tablets—the proposed algorithm can be deployed in point-of-care settings, including at-home monitoring. This capability significantly reduces unnecessary hospital visits while increasing maternal comfort and accessibility to reliable fetal health assessments.

In this work, we introduce Contrastive Ultrasound Video Representation Learning (CURL), a novel self-supervised framework for FM detection. CURL leverages a dual-contrastive loss—incorporating both spatial and temporal contrastive learning—to derive robust motion representations from extended ultrasound recordings. Furthermore, we propose a task-specific sampling strategy that effectively separates movement from non-movement segments, enabling flexible inference on arbitrarily long recordings through probabilistic fine-tuning.Our contributions pave the way for reliable, objective FM analysis that can be seamlessly integrated into clinical workflows and deployed in remote or mobile health settings with limited specialist availability.

\begin{figure*}[!ht]
	\centering
	\includegraphics[width=\textwidth]{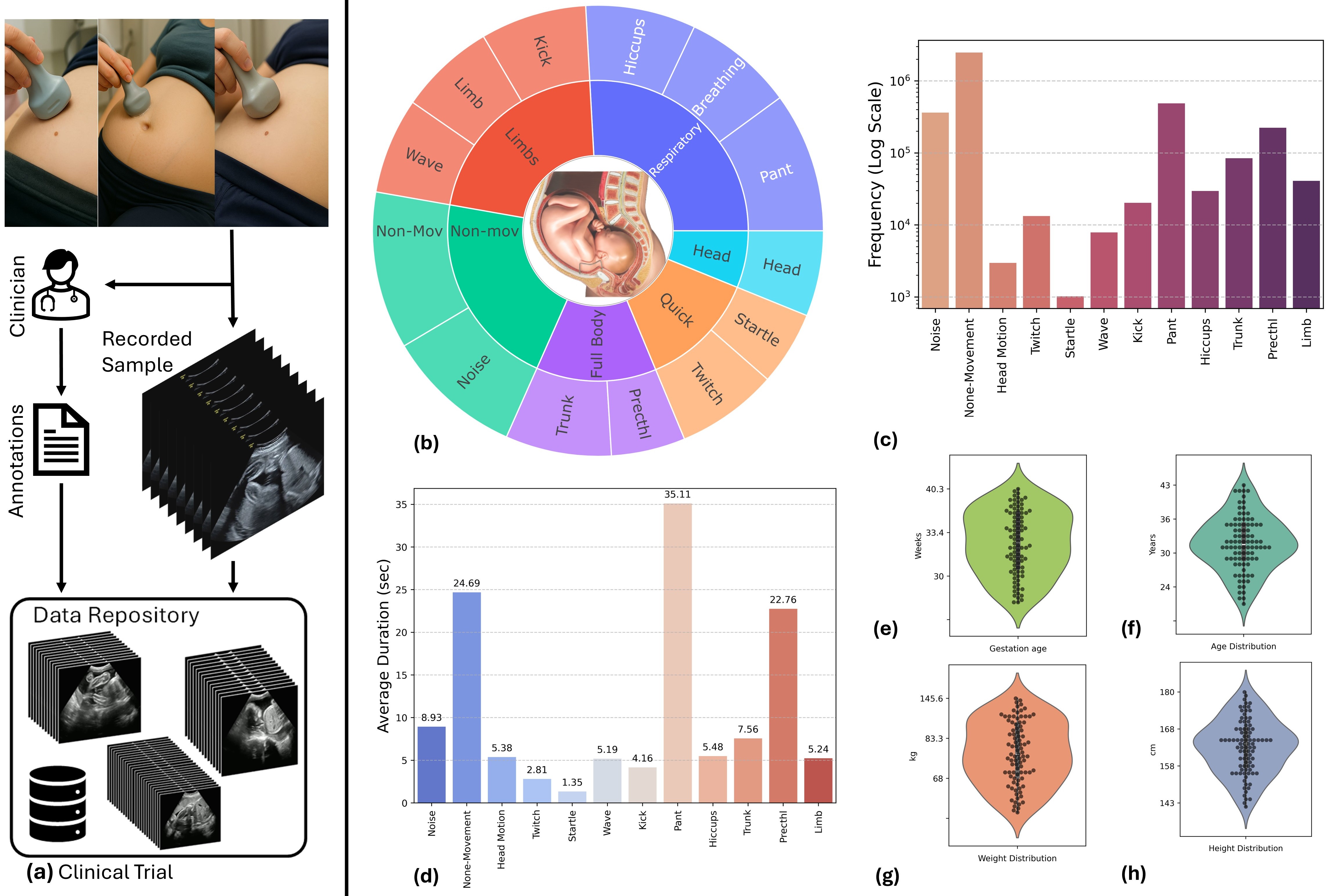}
	\caption{ Overview of the clinical trial workflow and dataset characteristics. Left panel (a) illustrates the data acquisition process, where trial data are collected, annotated and systematically stored in a centralized repository. Right panels (b)–(d) detail the movement features: (b) categorizes fetal movements as annotated by clinical experts, grouping them into super-categories based on motion type (e.g., breathing, panting, and hiccups are classified under respiratory movements); (c) presents the distribution of movement sub-classes on a logarithmic scale; and (d) displays the average duration (in seconds) for each movement sub-type. Panels (e)–(h) summarize the demographic data from clinical trial participants, with (e) showing gestational age (weeks), (f) depicting maternal age (years), (g) indicating maternal weight (kg), and (h) representing maternal height (cm). }
	\label{fig1}
\end{figure*}

\section{Literature Review}

\subsection{Sensor Based Approaches to Fetal Health Monitoring}

A variety of sensor-based methodologies have been investigated for monitoring fetal movements (FM), including both accelerometer-based detection systems and heart-rate based techniques. While these approaches have provided valuable insights into fetal activity, each is accompanied by limitations that have hindered widespread clinical adoption \cite{lai2018performance,alfirevic2017continuous}.

\subsubsection{Accelerometer-Based FM Monitoring}
Accelerometer-based systems capture FM by detecting abdominal vibrations. Nishihara et al. \cite{nishihara2008long} utilized capacitive accelerometers to measure FM, reporting a prevalence-adjusted bias-adjusted kappa (PABAK) of 0.75 relative to maternal perception. Ryo et al. \cite{ryo2012new} further validated this approach by comparing accelerometer outputs with concurrent ultrasound imaging, achieving a PABAK of 0.79 for gross movements, though only 0.36 for isolated limb actions.

Subsequent research has enhanced performance by employing multiple sensors and advanced signal processing techniques. Boashash et al. \cite{boashash2014passive} integrated four accelerometers with time-frequency analysis to reach a sensitivity of 0.78 and a precision of 0.83. More recently, Mesbah et al. \cite{mesbah2021automatic} combined wavelet transforms with machine learning, achieving binary classification accuracies between 0.87 and 0.95 on curated datasets. Despite these advances, most studies have been confined to controlled experimental environments, and challenges such as signal noise and interference from maternal activity still limit real-world applicability.

To bridge this gap, wearable sensor systems have been explored. For instance, Delay et al. \cite{delay2021novel} embedded an accelerometer in a wearable garment and, when compared against ultrasound, achieved a sensitivity of 0.81 and a precision of 0.77 for detecting fetal limb movements. Likewise, Lai et al. \cite{lai2018performance} evaluated acoustic sensors for FM detection, noting that a non-wearable setup attained a sensitivity of 0.78 relative to ultrasound, while a wearable version showed improved performance (sensitivity = 0.83, precision = 0.54) relative to maternal perception. Nevertheless, accelerometer-based approaches continue to be challenged by issues such as signal noise and the difficulty of isolating fetal signals from maternal movement.

\subsubsection{Heart-Rate–Based Approaches to FM Monitoring}

Heart-rate-based methods, predominantly cardiotocography (CTG) and fetal electrocardiography (fECG), indirectly infer fetal movements through characteristic variations in fetal heart rate (fHR) \cite{nageotte_2015}. CTG, the standard clinical method, identifies FM episodes from fHR accelerations, typically monitored via Doppler ultrasound \cite{marzbanrad2018cardiotocography}. Although widely available and continuous, CTG-based assessments remain indirect, with fHR changes influenced by factors unrelated to movement, such as maternal physiological states and uterine activity \cite{rooijakkers2014fetal}.

Fetal ECG, obtained from maternal abdominal electrodes, offers higher temporal resolution than CTG and enables detailed analysis of cardiac waveform morphology \cite{nageotte_2015}. Machine-learning models have enhanced movement detection by extracting transient amplitude and morphological features from fetal QRS complexes \cite{liu2014multi,rooijakkers2015feasibility}. For example, dilated convolutional neural networks isolate QRS complexes from noisy abdominal recordings, and dual-path LSTM architectures achieve F1 scores of up to 95.3\% in QRS detection by modeling transient amplitude fluctuations associated with movement \cite{shokouhmand2022fetal}. Simplified single-lead ECG approaches have also shown clinically relevant performance (sensitivity 0.67, specificity 0.90), facilitating more accessible ambulatory monitoring \cite{rooijakkers2015feasibility,rooijakkers2014fetal}. However, separating subtle fetal signals from maternal cardiopulmonary artifacts remains challenging.

Fetal magnetocardiography (fMCG) provides research-grade fidelity with excellent temporal and spatial resolution by measuring magnetic fields associated with fetal cardiac activity \cite{govindan2011novel,avci2017tracking,lutter2011indices}. Despite its accuracy, the high cost, bulky equipment, and specialized infrastructure required restrict fMCG to a few research centers \cite{stinstra2002influence}.

Overall, while accelerometer, CTG, and fECG methodologies have improved fetal movement assessment, their indirect detection mechanisms, susceptibility to noise, and practical limitations underscore the need for alternative, direct measurement approaches. These challenges motivate the continued development of video-based ultrasound methods capable of capturing detailed, direct spatiotemporal fetal movement patterns.

\subsection{AI-Enhanced Ultrasound and Imaging Technologies}
The integration of AI into fetal ultrasound analysis has revolutionized the detection, classification, and interpretation of fetal anatomical structures. Traditional ultrasound imaging remains central to fetal assessment, providing essential biometric measurements such as head circumference and limb length, as well as detailed visualization of fetal anatomy \cite{yaqub2012automatic,sobhaninia2019fetal,lei2014automatic,yu2016fetal}. However, AI-driven techniques have expanded these capabilities by automating diagnostic processes and enhancing image interpretation.

Early applications of AI in fetal imaging focused on classifying static ultrasound images. For example, Ishikawa et al. \cite{ishikawa2019detecting} developed a framework for recognizing and classifying fetal parts—such as the head, trunk, and limbs—to accurately predict fetal positioning. Recent studies further highlight the promise of AI-enhanced ultrasound for fetal health assessment. Heuvel et al. \cite{van2019automated} introduced a deep learning model for measuring head circumference, particularly useful in resource-limited settings. Dozen et al. \cite{dozen2020image} applied a time-series-based deep learning network to segment dynamically changing fetal heart structures, while Ravishankar et al. \cite{ravishankar2016hybrid} combined traditional tissue recognition with AI to improve abdominal region detection in two-dimensional ultrasound images.

Furthermore, advancements in AI-driven fetal echocardiography illustrate the transformative potential of these technologies. Chen et al. \cite{chen2020automatic} devised a model for quantifying fetal heart ventricles, and Arnaout et al. \cite{arnaout2021ensemble} implemented a neural network-based approach for early detection of congenital heart diseases. These techniques collectively demonstrate how deep learning can reduce operator dependency while enhancing diagnostic accuracy.

Despite these significant advances, most AI-enhanced ultrasound methods have primarily focused on static images, limiting their ability to capture the continuous and dynamically evolving nature of fetal movements. This limitation underscores the need for integrating video-based analysis with AI to enable comprehensive, temporally-informed fetal monitoring. By capturing both spatial and temporal information, video-based approaches promise to provide a more complete and nuanced assessment of fetal health \cite{stanojevic2023fetal,zhao2023lower}, addressing challenges that static imaging alone cannot overcome.

\subsection{Potential of Video Analysis for Fetal Monitoring}
Traditional ultrasound imaging techniques, while pivotal in fetal health assessment, are fundamentally limited by their focus on static images rather than continuous, dynamic monitoring. Many conventional approaches analyze individual frames, which restricts their capacity to concurrently track multiple evolving parameters and capture the full spectrum of fetal movements over time. Only a few studies have attempted to harness the temporal information available in video data \cite{inubashiri2021new,dandil2021fetal,turkan2025fetalmovnet,turkan2023automatic}, though these early efforts are constrained by several limitations.

Early work \cite{inubashiri2021new} in this domain leveraged 4D fetal ultrasound data to detect general movements (GM) using optical flow analysis. In these studies, researchers analyzed short (10-second) 4D ultrasound clips to quantify movement velocities—such as maximum, median, average, and mode velocities—by computing motion vectors. Although this approach provided a quantitative measure of fetal activity, it was hampered by significant challenges. Notably, the inability to isolate the fetus from surrounding structures (e.g., the umbilical cord and placenta) compromised the accuracy of the measurements. Moreover, strict adherence to the ALARA (As Low As Reasonably Achievable) principle limited both the ultrasound exposure and frame rate (often to only 10 frames per second), which in turn risked overlooking subtle movement nuances.

Subsequent efforts, such as those by Dandil et al. \cite{dandil2021fetal}, employed a frame-by-frame processing strategy using the YOLOv5 network to identify fetal anatomical landmarks in ultrasound videos. By tracking the 2D trajectories of landmarks—primarily for the fetal head and body—they aimed to visualize movement patterns. Their extension using LSTM \cite{turkan2023automatic} 
 networks to model temporal dynamics further refined the extraction of 2D motion trajectories. However, these methods depend heavily on the accuracy of frame-by-frame landmark detection and the subjective interpretation of the resulting trajectories, limiting their capacity to fully characterize the sequential evolution of fetal movements.

More recently, Turkan et al. \cite{turkan2025fetalmovnet} introduced FetalMovNet that integrates convolutional neural networks (CNNs) with an attention mechanism to capture spatio-temporal patterns for fetal movement classification. Designed to process fixed-length input sequences (typically 10 frames capturing a single movement type), this approach is constrained by its rigid temporal window. Such constraints make it less effective for real-life applications where the onset and duration of fetal movements vary considerably, and where continuous monitoring over extended periods is essential. Additionally, the inability to process arbitrarily long videos curtails its utility in both clinical settings and at-home monitoring scenarios.

In contrast, our proposed Contrastive Ultrasound Video Representation Learning (CURL) framework is specifically designed to overcome these challenges. By employing a dual-contrastive loss that simultaneously optimizes spatial and temporal features, CURL effectively captures the nuanced dynamics of fetal movements from extended ultrasound video streams. Our framework introduces two complementary sampling strategies: a clean-cut sampling method during the self-supervised representation learning phase and a sliding window sampling strategy for practical, real-time inference on arbitrarily long videos. This dual approach enhances movement detection robustness and aligns with standard practices in video action recognition, rendering our model highly adaptable for both clinical and home monitoring applications.

Ultimately, by enabling the continuous capture and analysis of ultrasound video streams, video-based fetal monitoring offers a transformative opportunity to improve prenatal care. The ability to track temporal dynamics facilitates the simultaneous evaluation of multiple fetal parameters, thereby enhancing diagnostic accuracy and clinical decision-making, and paving the way for more objective, automated and real-time insights into fetal development and well-being.
\begin{table}[t]
\centering
\caption{Eligibility Criteria for Participant Selection}
\label{tab:selection_criteria}
\renewcommand{\arraystretch}{1.4}
\setlength{\tabcolsep}{6pt} 
\begin{tabular}{|p{0.45\columnwidth}|p{0.45\columnwidth}|}
\toprule
\textbf{Inclusion Criteria} & \textbf{Exclusion Criteria} \\
\midrule
\textbullet\ Age $\geq$ 18 years & \textbullet\ Allergy to latex or elastoplast \\
\textbullet\ Singleton pregnancy & \textbullet\ Presence of fetal congenital abnormality \\
\textbullet\ Gestational age $\geq$ 28 weeks & \textbullet\ High dependency on medical care \\
\textbullet\ Awareness of normal fetal movements on the day of ultrasound & \textbullet\ Maternal weight $>$ 200 kg \\
 & \textbullet\ Existing relationship with the research team \\
 & \textbullet\ Special circumstances at the discretion of the research midwife (e.g., previous stillbirth, severe intrauterine growth restriction (IUGR), fetal abnormalities requiring further assessment, social or mental health concerns) \\
\bottomrule
\end{tabular}
\end{table}

\begin{table}[]
\centering
\caption{Definitions of Fetal Movement Annotations}
\label{tab:fetal_movements}
\renewcommand{\arraystretch}{1.3}
\setlength{\tabcolsep}{4pt} 
\begin{tabular}{|p{0.3\columnwidth}|p{0.65\columnwidth}|}
\hline
\textbf{Movement Type} & \textbf{Definition} \\
\hline
Prechtl Movements & Gross body movements lasting seconds to minutes, involving variable sequences of arm, leg, neck, and trunk motion with gradual onset and offset. \\
Head Motion & Movements including extension, flexion, and lateral turning of the head. \\
Twitch & Brief limb jerk ($\leq 1$ sec) with minimal displacement, without rhythmic pattern. \\
Startle & Abrupt, shock-like whole-body jerk, lasting $\sim$1 sec. \\
Wave & Slow upper limb movement ($\geq 1$ sec) with no rhythm or pattern. \\
Kick & Slow lower limb movement ($\geq 1$ sec) with no rhythm or pattern. \\
Breathing & Slow, irregular diaphragm contractions, causing outward or downward displacement of the abdomen. \\
Panting & Rapid mild diaphragm contractions, occurring regularly or irregularly. \\
Hiccups & Brief, uniform thoracic movements, sometimes affecting limb motion. \\
Trunk Roll & Isolated rolling motion of the fetal trunk. \\
\hline
\end{tabular}
\end{table}

\section{Dataset Collection Protocol and Characteristics}

We collected 92 fetal ultrasound video recordings (30 min each) from two clinical sites: the Department of Obstetrics \& Gynaecology, Monash University, Clayton, VIC 3168, and the Division of Perinatal Medicine, Kolling Institute, The University of Sydney, St Leonards, NSW 2065. Participants were enrolled under strict inclusion/exclusion criteria to ensure clinical relevance and data quality. All sessions followed a unified acquisition protocol to guarantee consistency, reliability, and reproducibility across the dataset.

\subsection{Ethical Considerations}

The study protocol was approved by Monash Health’s Human Research Ethics Committee (RES-17-0000-028) and registered with the Australian New Zealand Clinical Trials Registry (ACTRN12617000410358).

\subsection{Participant Selection Criteria}

Participants were recruited based on well-defined inclusion and exclusion criteria to ensure data consistency and minimize confounding variables. The selection process was carried out by trained research midwives in collaboration with the hospital’s antenatal unit. The eligibility criteria are summarized in Table~\ref{tab:selection_criteria}.

\begin{figure*}[!t]
	\centering
	\includegraphics[width=\textwidth]{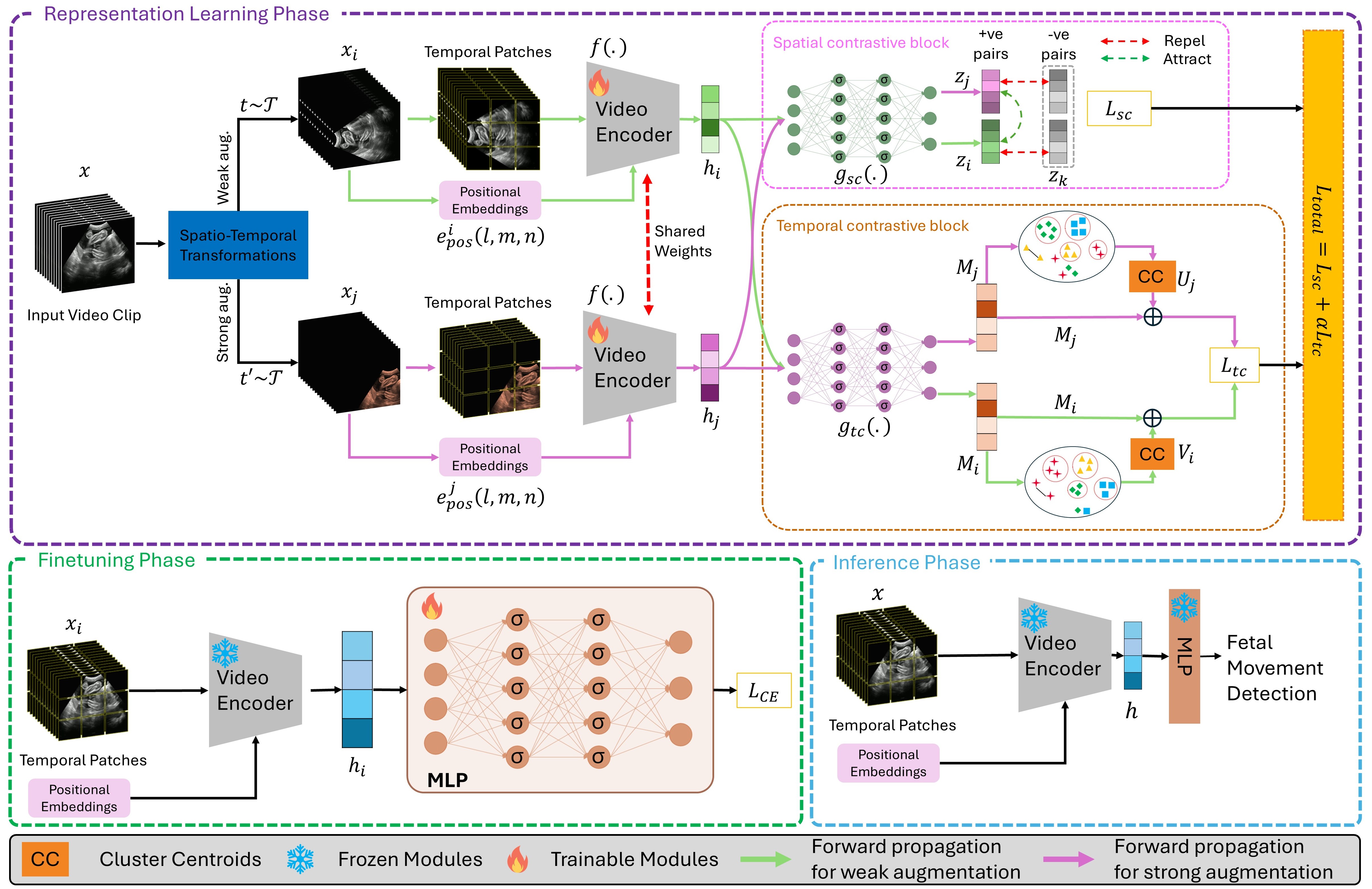}
	\caption{Schematic overview of the proposed CURL framework pipeline, comprising six sequential stages. (1) Video clip extraction is performed using either a clean-cut or sliding-window sampling strategy. (2) Spatio-temporal augmentations are applied to construct paired video clips \((x_i, x_j)\), which are then passed through a shared video encoder \(f(\cdot)\) to obtain corresponding feature representations \((h_i, h_j)\). These representations are subsequently routed into two distinct branches. (3) In the spatial contrastive learning branch, an MLP projection head \(g_{sc}(\cdot)\) projects the features into a latent space for instance-level spatial contrastive learning. (4) In parallel, the temporal contrastive learning branch uses another MLP projection head \(g_{tc}(\cdot)\) to map the features into a space that facilitates the clustering of temporally consistent motion patterns. (5) A fine-tuning phase follows, wherein the framework is adapted for the downstream task via a task-specific head. (6) Finally, during inference, the model processes input clips in a sliding-window fashion using a single forward pass.}
	\label{fig1}
\end{figure*}

\subsection{Participant Recruitment and Data Collection Process}

The recruitment process followed a structured approach to ensure voluntary participation while maintaining confidentiality and adherence to ethical standards.

\textbf{Screening for Eligibility:} Potential participants were identified using the hospital’s \textit{Topaz Antenatal Day Clinic List}. The \textit{Birthing Outcomes System (BOS)} and Electronic Medical Records (EMR) were reviewed to assess suitability. The \textit{Fetal Kicks Participant Database} was consulted to confirm that the patient had not been approached previously.

\textbf{Initial Approach and Consent:} Participants who met the eligibility criteria were approached by a research midwife, who provided a detailed explanation of the study. If the participant expressed interest, they received an information sheet and a consent form for review. Upon obtaining written consent, a suitable ultrasound appointment was scheduled.

\textbf{Pre-Ultrasound Preparation:} On the day of data collection, the participant was accompanied to the ultrasound room while adhering to necessary hospital safety protocols. A unique participant identifier (\textbf{FK-***}) was assigned for anonymization. Before the ultrasound, the participant completed a pre-ultrasound questionnaire, and the research midwife recorded maternal demographics and antenatal risk factors.

\textbf{Handling Declined Participation:} If a participant chose not to participate, they were thanked for their time, reassured that their decision would not affect their medical care, and their non-participation was documented in the \textit{Fetal Kicks Participant Database} to prevent further contact.

\subsection{Dataset Characteristics}

The dataset comprises of 92 ultrasound video recordings collected during routine antenatal checkups, each spanning 30 minutes in duration. The videos were recorded at an original resolution of 976×736 pixels with a frame rate of 23 frames per second (FPS). To facilitate movement analysis, video segments were extracted at 10 FPS, generating 50 frames per 5-second clip for subsequent processing.

\subsubsection{Data Processing and Noise Removal}

To ensure high-quality movement detection, a rigorous pre-processing pipeline was implemented to eliminate non-clinical noise and artifacts. At the start and end of each recording, six distinct rubber-hammer taps were applied as temporal markers, clearly delineating the boundaries of the ultrasound session. These markers enabled the precise removal of extraneous segments; specifically, the first and last two minutes of each video were excised, thereby ensuring that only clinically relevant frames were retained for subsequent analysis.

In addition, we explored various segmentation strategies to optimize the extraction of motion features. For the self-supervised representation learning phase, a clean-cut sampling method was employed to extract distinct, non-overlapping segments. For fine-tuning and real-time analysis, sliding-window sampling strategy was adopted to capture continuous temporal dynamics. A comprehensive discussion of these approaches and their impact on performance is provided in the Methods section and ablation studies.

\subsubsection{Categories of Movement and Non-Movement}
Each fetal movement type was defined based on clinical motion patterns. The dataset is annotated into two primary categories: non-movement (including noise) and movement.Detailed subcategories are provided in Table~\ref{tab:fetal_movements}.

\subsubsection{External and Probe-Induced Motion}

External fetal movements may arise from maternal actions such as coughing or laughing, leading to uterine contractions. Additionally, probe movements made by the sonographer to optimize fetal imaging may introduce external disturbances, which were separately categorized to ensure accurate fetal motion classification.

\section{Methodology}

In our approach, we begin by training a video encoder in a self-supervised manner using the proposed Contrastive Ultrasound Video Representation Learning (CURL) framework, see Figure \ref{fig1}. For every input video clip \( x \), two distinct sets of transformations, \(\mathcal{T}\) and \(\mathcal{T}'\), are applied to generate two correlated views: $ x_i = \mathcal{T}(x) \quad \text{and} \quad x_j = \mathcal{T}'(x)$.
These augmented samples are then passed through a shared backbone \( f(\cdot) \) to extract feature representations \( h_i \) and \( h_j \). Subsequently, these representations are fed into two separate MLP projection heads, \( g_{sc}(\cdot) \) and \( g_{tc}(\cdot) \), corresponding to spatial and temporal contrastive learning branches, respectively. This dual contrastive loss strategy ensures that our model learns robust representations sensitive to both fine-grained anatomical details and the dynamic progression of fetal movements.

\begin{figure}[]
	\centering
	\includegraphics[width=3in]{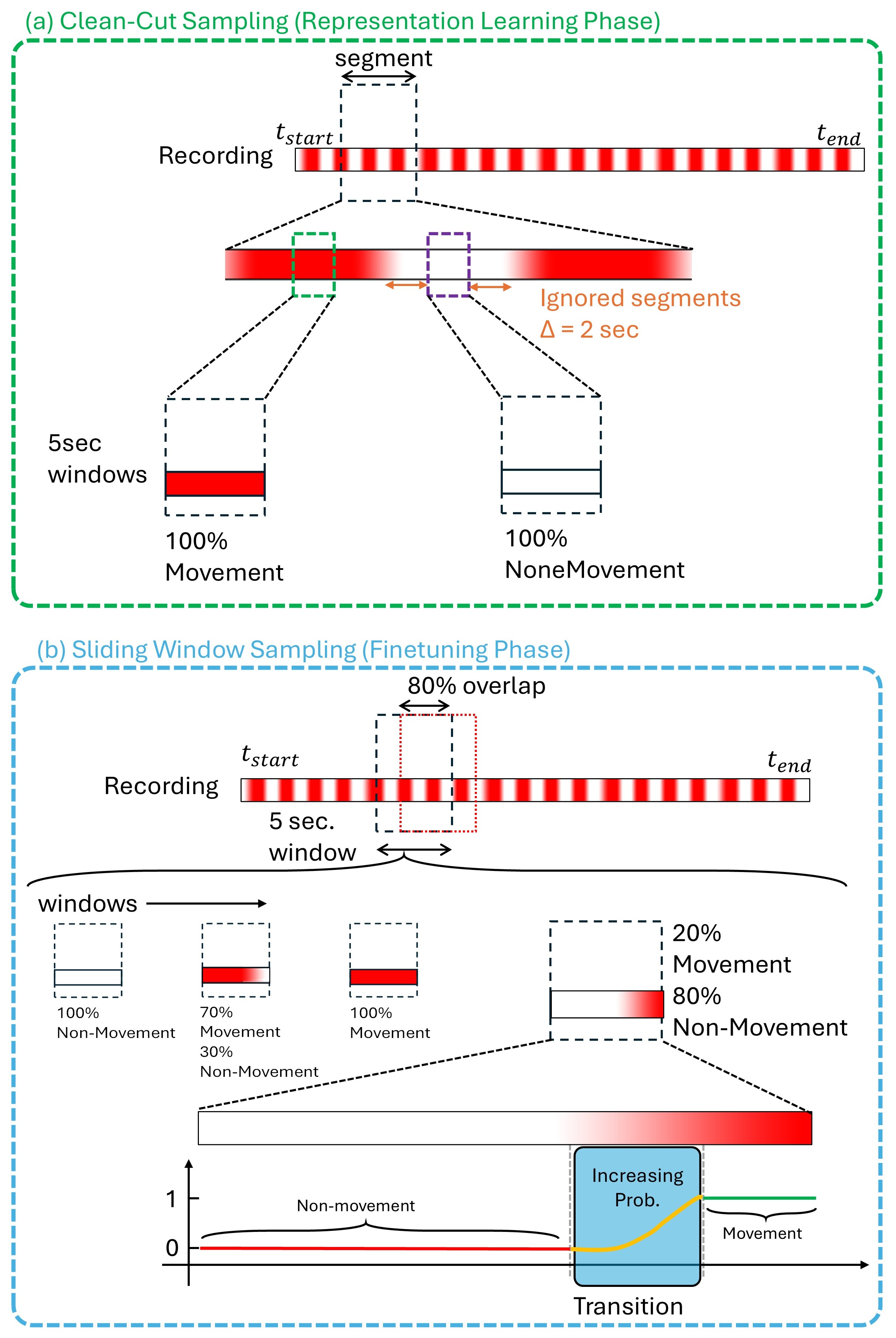}
	\caption{Comparative visualization of key characteristics of clean-cut sampling versus sliding-window sampling. (a)Clean-cut sampling selects distinct, non-overlapping video segments, thereby providing independent samples that capture broader temporal contexts.(b) Sliding-window sampling continuously moves a fixed-size window over the video, generating overlapping segments that offer finer-grained temporal dynamics at the expense of some redundancy. }
	\label{sampling2}
\end{figure}

\subsection{Video Encoder}

Our framework employs the Vision Transformer (ViT) architecture, building upon the Masked Autoencoders (MAE) paradigm introduced by He et al. \cite{heMaskedAutoencodersAre2021} and its extension to spatiotemporal data by Feichtenhofer et al. \cite{feichtenhoferMaskedAutoencodersSpatiotemporal2022}. We adopt the official implementation of the Spatiotemporal MAE (MAE-ST) model\cite{feichtenhoferMaskedAutoencodersSpatiotemporal2022}, incorporating specific adaptations to suit our fetal movement analysis objectives.

Let the input video clip be denoted as $\mathbf{x} \in \mathbb{R}^{C \times T \times H \times W}$, where $C$ represents the number of channels, $T$ the temporal length (number of frames), and $H$ and $W$ the spatial dimensions. The video is partitioned into non-overlapping 3D patches of size $(t, h, w) = (2, 16, 16)$, resulting in $N = \frac{T}{t} \times \frac{H}{h} \times \frac{W}{w}$ total patches. For instance, with $T = 50$ frames and spatial dimensions $224 \times 224$, we obtain $25 \times 14 \times 14$ patches.

Each 3D patch is flattened and projected into a fixed-dimensional embedding space via a linear layer. To preserve spatiotemporal information, we incorporate separable positional embeddings: a temporal embedding $\mathbf{e}_{\text{temp}} \in \mathbb{R}^{\frac{T}{t} \times D}$ and a spatial embedding $\mathbf{e}_{\text{spat}} \in \mathbb{R}^{\left(\frac{H}{h} \times \frac{W}{w}\right) \times D}$, where $D$ is the embedding dimension. The combined positional embedding for a patch at temporal index $l$ and spatial indices $(m, n)$ is given by: $ \mathbf{e}_{\text{pos}}(l, m, n) = \mathbf{e}_{\text{temp}}(l) + \mathbf{e}_{\text{spat}}(m, n)$. This positional embedding is added to the patch embedding to form the input token:$\mathbf{z}_{l, m, n}^{\text{in}} = \mathbf{z}_{l, m, n} + \mathbf{e}_{\text{pos}}(l, m, n)$. The sequence of tokens is then flattened and processed by the ViT encoder, which comprises 12 transformer blocks, each with 12 attention heads and an embedding dimension of 768. Each block includes multi-head self-attention mechanisms and feed-forward networks, facilitating the modeling of complex spatiotemporal dependencies \cite{sun2025masked}.

While our implementation closely follows the MAE-ST architecture, we introduce specific modifications to tailor it for fetal movement analysis:  First, we extract 5-second clips at 10 FPS—corresponding to the median sub-movement duration of 5 s observed in our data (e.g., head motion 5.38 s, wave 5.19 s, limb movement 5.24 s; Figure \ref{fig1})—thereby ensuring that most movement types are fully captured while avoiding unnecessary temporal redundancy. Second, we replace the conventional positional embedding with separable embeddings that independently encode temporal and spatial dimensions, which enhances the model’s ability to learn spatiotemporal features relevant to movement patterns. Finally, after pre-training the encoder on large-scale data, we perform a dedicated fine-tuning step using our fetal movement dataset, directly optimizing the model for the tasks of movement detection and classification. Together, these modifications enable the model to effectively learn representations pertinent to fetal movement analysis, leveraging the strengths of the ViT architecture in capturing spatiotemporal patterns.

\subsection{Contrastive Ultrasound Video Representation Learning (CURL) Framework}

Our CURL framework leverages self-supervised contrastive learning to extract robust video representations from fetal ultrasound recordings. Unlike conventional image-based approaches, videos inherently include a temporal dimension that is critical for capturing the dynamic evolution of fetal movements. To address this, our framework integrates dual contrastive losses—one focused on spatial features $L_{sc}$ and the other on temporal dynamics $L_{tc}$—to ensure that both anatomical details and motion patterns are robustly encoded for downstream detection tasks.

\subsubsection{Spatial Contrastive Learning}

Spatial contrastive learning in CURL is designed to emphasize discriminative anatomical features—such as limb and head structures—that are indicative of fetal movement. Following a paradigm similar to SimCLR \cite{chen2020simple}, we randomly sample a mini-batch of $N$ images and apply two distinct data augmentations to each image, resulting in $2N$ augmented samples. These augmentations simulate variations in appearance while preserving key ultrasound features.

We then compute contrastive loss using the InfoNCE formulation \cite{oord2018representation}. For a given augmented sample \( x_i \) with corresponding positive pair \( x_j \), the loss is defined as:

\begin{equation}
\ell_i = -\log \frac{\exp\big(\text{sim}(z_i, z_j)/\tau_{\text{ins}}\big)}{\sum_{k=1,\, k\neq i}^{2N} \exp\big(\text{sim}(z_i, z_k)/\tau_{\text{ins}}\big)},
\end{equation}

where \(z_i\) and \(z_j\) are the projected features, \(\text{sim}(\cdot,\cdot)\) denotes cosine similarity, and ($\tau_{ins}$) is the temperature parameter, set to 0.1 by default following \cite{chen2020simple}. This objective encourages the model to bring representations of the same clip closer while pushing apart others in the batch, thereby focusing on consistent anatomical details essential for fetal movement identification.

\begin{figure}[]
	\centering
	\includegraphics[width=3.5in]{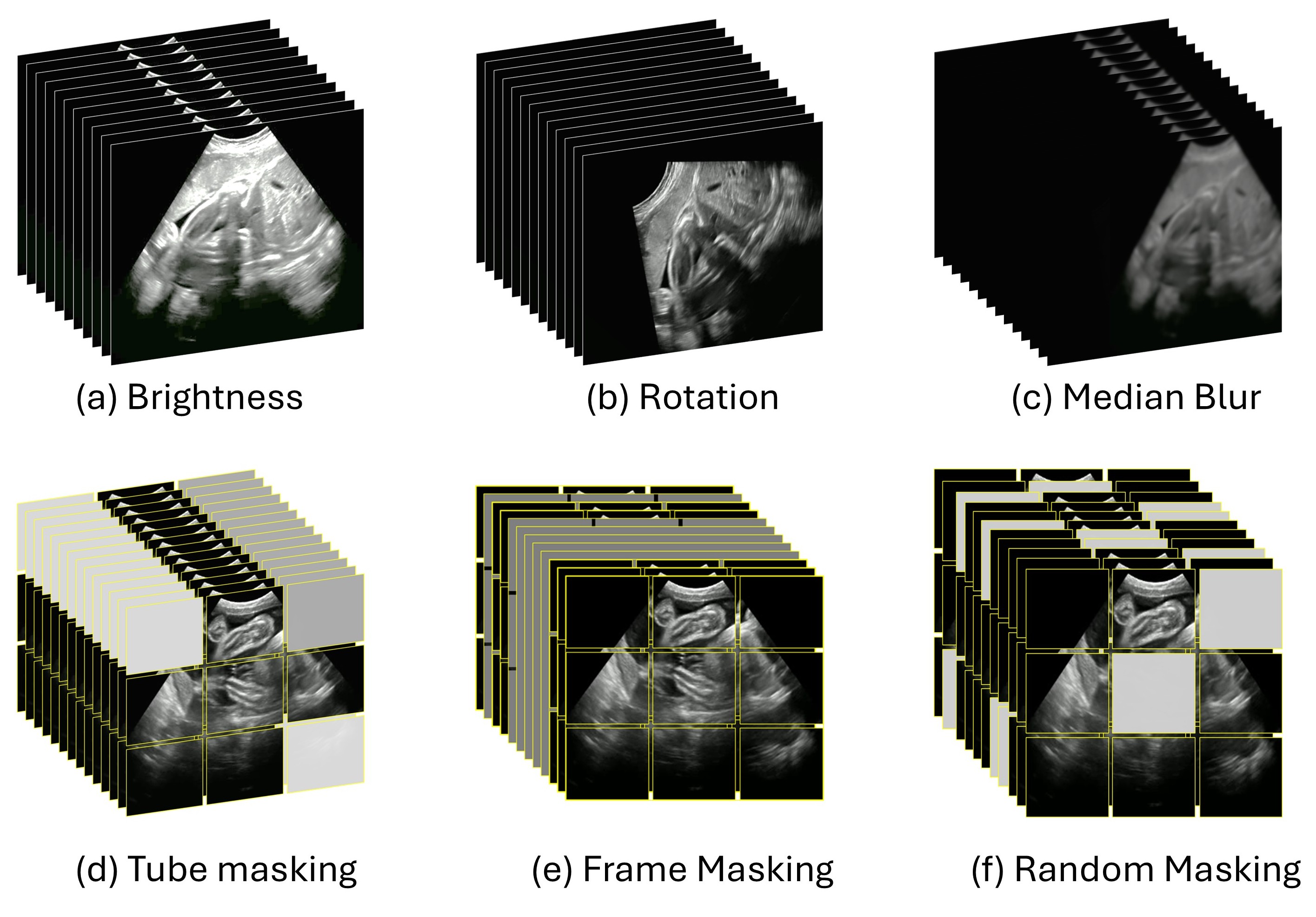}
	\caption{Example video clips demonstrating spatial and temporal augmentation techniques. Panels (a)–(c) illustrate spatial augmentations applied to video frames: (a) brightness adjustment, (b) rotation, and (c) median blur filtering. Panels (d)–(f) display temporal augmentations: (d) tube masking, (e) frame masking, and (f) random masking.}
	\label{aug_imgs}
\end{figure}

\subsubsection{Temporal Contrastive Learning}

Temporal dynamics play a crucial role in capturing subtle variations of fetal movements over time. Directly applying a contrastive loss to temporally adjacent frames, however, is problematic due to their inherent correlation, as such frames might erroneously be considered negatives. To address this challenge, we propose a clustering-based temporal contrastive strategy.

Two augmented views of a video segment are processed through a shared backbone \( f(\cdot) \) to generate temporally enriched feature representations \( h_i \) and \( h_j \). These representations are then projected via a temporal-specific MLP head \( g_{tc}(\cdot) \), yielding temporally aware embeddings \( M_i \) and \( M_j \).

Within each mini-batch, we perform \( K \)-means clustering on embeddings to group them into semantically coherent clusters representing similar movement patterns, such as breathing, kicking, or brief movements like twitches or startles. For clarity, the centroids for embeddings from the two augmented views are denoted by \( U = \{u_1, u_2, \dots, u_K\} \) and \( V = \{v_1, v_2, \dots, v_K\} \), respectively, and each embedding is assigned to its closest centroid. Instead of solely contrasting individual instances, our method promotes embeddings from one view to predict the corresponding cluster centroid from the other view.  

The temporal contrastive loss (cluster-aware loss) for an embedding \( M_i \) assigned to centroid \( v_i \in V \) is defined as:

\begin{equation}
\hat{\ell}_i = -\log \frac{\exp\big(\text{sim}(M_i, v_i)/\tau_{\text{ca}}\big)}{\sum_{k=1,\, k\neq i}^{K} \exp\big(\text{sim}(M_i, v_k)/\tau_{\text{ca}}\big)},
\end{equation}

with analogous formulation for the pared embedding \( M_j \) with its corresponding centroid \( u_i \in U \), and \(\tau_{\text{ca}}\) is the temperature parameter controlling similarity scaling, set to 0.5 following prior literature on clustering-based contrastive learning \cite{zhang2021supporting}. This formulation encourages temporal embeddings to consistently represent semantically similar movement segments across different augmented views.

To prevent trivial solutions where embeddings collapse into a few dominant clusters, we introduce cluster entropy regularization inspired by \cite{chakraborty2020entropy}. Specifically, we define a joint probability matrix $P$ over cluster assignments across embeddings from both views, computed as $P = \frac{1}{N}(M \cdot V^\top)(M' \cdot U^\top)^\top$. Using $P$, we calculate the cluster entropy (mutual information) as follows:

\begin{equation}
I(U,V) = \sum_{i=1}^{K}\sum_{j=1}^{K}P_{ij}\ln\frac{P_{ij}}{P_i P_j}
\end{equation}

where $P_{ij}$ denotes the joint probability of co-assignment to centroids $u_i$ and $v_j$, and $P_i$, $P_j$ are marginal probabilities of individual clusters. The final temporal contrastive loss integrates this entropy-based regularization term:

\begin{equation}
L_{\text{tc}} = \frac{1}{N}\sum_{i=1}^{N}\hat{\ell}_i - I(U,V)
\end{equation}

We set the number of clusters \( K \) empirically to 10, informed by preliminary analyses demonstrating that this setting captures the intrinsic diversity of fetal movement types present in our dataset without incurring instability or cluster redundancy.

Through joint optimization of the spatial and temporal contrastive losses, CURL effectively learns representations sensitive to both fine-grained anatomical features and the dynamic evolution of fetal movements. This dual-contrastive approach supports robust downstream fetal movement detection and classification tasks from ultrasound recordings.

\begin{table*}[t]
\centering
\caption{Performance comparison of fetal movement detection approaches on ultrasound dataset. We compare supervised single-stage methods with two-stage self-supervised approaches. Best results in \textbf{bold}, second best underlined. Reported values represent the mean $\pm$ standard deviation.}
\label{tab:benchmark}
\small
\setlength{\tabcolsep}{4pt} 
\begin{tabular}{@{} ll *{6}{S[table-format=2.2(2)]} @{}}
\toprule
\textbf{Method} & \textbf{Backbone} & \textbf{Spec.} & \textbf{Sen.} & \textbf{W.Prec} & \textbf{W.F1} & \textbf{bACC} & \textbf{AUROC} \\
 & & {\scriptsize(Rec. Non-Mov)} & {\scriptsize(Rec. Mov)} & & & & \\
\midrule
\multicolumn{8}{@{}l}{\textit{Supervised Single-Stage Methods}} \\
\cmidrule(lr){1-8}
CNN+LSTM\cite{carreira2017quo}           & CNN+LSTM  & 55.10 \pm 1.56 & 36.62 \pm 1.18 & 46.84 \pm 2.58 & 44.42 \pm 2.80 & 45.68 \pm 2.02 & 43.16 \pm 2.63 \\
FetalMovNet\cite{turkan2025fetalmovnet} & Custom    & 61.22 \pm 2.07 & 63.06 \pm 1.65 & 60.08 \pm 1.53 & 59.37 \pm 1.89 & 58.71 \pm 1.15 & 60.71 \pm 2.35 \\
I3D\cite{carreira2017quo}               & I3D       & 49.06 \pm 2.55 & 62.61 \pm 1.76 & 49.34 \pm 1.92 & 56.01 \pm 2.32 & 52.35 \pm 1.04 & 59.67 \pm 2.20 \\
TwoStream I3D\cite{carreira2017quo}      & TwoStream & 59.41 \pm 1.52 & 64.36 \pm 1.44 & 61.89 \pm 1.39 & 63.12 \pm 1.89 & 61.88 \pm 2.27 & 64.36 \pm 2.19 \\
SlowFast\cite{feichtenhofer2019slowfast} & R101+NL   & 64.36 \pm 2.89 & 69.31 \pm 2.54 & 66.84 \pm 1.20 & 68.07 \pm 2.11 & 66.83 \pm 1.97 & 69.31 \pm 2.57 \\
X3D-XXL\cite{feichtenhofer2020x3d}      & X3D       & 62.38 \pm 1.88 & 67.33 \pm 1.79 & 64.86 \pm 1.51 & 66.09 \pm 1.58 & 64.85 \pm 2.38 & 67.33 \pm 1.44 \\
MViTv2-L\cite{li2022mvitv2}              & MViTv2    & 66.67 \pm 2.35 & 71.43 \pm 2.47 & 69.05 \pm 1.51 & 70.24 \pm 1.64 & 69.05 \pm 2.73 & 71.43 \pm 1.59 \\
VideoSwin-L\cite{liu2022video}           & Swin-L    & 67.29 \pm 1.80 & 71.96 \pm 1.61 & 70.09 \pm 1.28 & 71.27 \pm 2.91 & 69.63 \pm 1.96 & 72.90 \pm 2.95 \\
ViViT-L\cite{arnab2021vivit}             & ViViT-L   & 68.52 \pm 1.64 & 74.07 \pm 2.84 & 71.30 \pm 1.40 & 72.69 \pm 2.01 & 73.30 \pm 1.74 & 74.07 \pm 1.50 \\

\midrule
\multicolumn{8}{@{}l}{\textit{Self-Supervised Two-Stage Approaches}} \\
\cmidrule(lr){1-8}
SimCLR Inflated\cite{chen2020simple}    & R3D-50    & 49.50 \pm 2.75 & 54.46 \pm 2.64 & 51.98 \pm 2.45 & 53.22 \pm 2.65 & 53.98 \pm 2.53 & 54.46 \pm 2.18 \\
MAE\cite{feichtenhofer2022masked}       & ViT-L     & 67.33 \pm 1.21 & 71.29 \pm 2.71 & 68.81 \pm 1.64 & 70.30 \pm 2.78 & 69.31 \pm 2.12 & 71.29 \pm 2.44 \\
CVRL\cite{qian2021spatiotemporal}       & SlowFast  & 65.35 \pm 1.56 & 69.31 \pm 1.01 & 67.33 \pm 2.45 & 68.32 \pm 1.73 & 67.33 \pm 1.13 & 69.31 \pm 2.21 \\
CVRL\cite{qian2021spatiotemporal}       & ViT-L     & 69.31 \pm 1.91 & 74.26 \pm 2.50 & 73.27 \pm 3.18 & 72.53 \pm 2.34 & 71.78 \pm 3.29 & 73.27 \pm 3.16 \\

\rowcolor{gray!10}
\textbf{CURL (Ours)}                     & SlowFast  & 72.28 \pm 1.47 & 75.25 \pm 1.35 & 74.26 \pm 1.67 & 75.00 \pm 1.19 & 73.76 \pm 1.38 & 76.24 \pm 1.14 \\
\rowcolor{gray!10}
\textbf{CURL (Ours)}                     & ViT-L     & {\textbf{75.97 $\pm$ 1.97}} & {\textbf{78.01 $\pm$ 1.12}} & {\textbf{75.23 $\pm$ 2.32}} & {\textbf{81.17 $\pm$ 1.96}} & {\textbf{80.74 $\pm$ 2.09}} & {\textbf{81.60 $\pm$ 2.81}} \\
\bottomrule
\end{tabular}

\vspace{0.5em}
\footnotesize
\textbf{Abbreviations}: Spec. = Specificity, Sen. = Sensitivity, W.Prec = Weighted Precision, W.F1 = Weighted F1-score, bACC = Balanced Accuracy
\end{table*}

\subsection{Downstream Adaptation}
\subsubsection{Informed Sampling Strategy}

Fetal ultrasound recordings naturally alternate between movement and non-movement intervals, and our objective is to ensure that learned representations for movement segments remain consistent regardless of their position in the video, while also being distinct from non-movement segments. To achieve this, we deploy two complementary sampling strategies tailored to the self-supervised learning and fine-tuning phases.

\paragraph{\textbf{Clean-cut sampling}} For the self-supervised phase, our goal is to generate clean, unambiguous clips that maximize the effectiveness of contrastive learning. We first partition the full video (of duration \(T\)) into disjoint segments that are exclusively characterized as either fetal movement or non-movement. This segmentation avoids transitional frames where the state shifts, thus preventing potential ambiguity during representation learning. We refer to this approach as \emph{clean-cut sampling}. Moreover, for each segment defined by its start \(t_{\text{start}}\) and end \(t_{\text{end}}\), we discard a fixed duration (\(\Delta\)) from both ends to eliminate boundary artifacts. Formally, the final clip \(C_s\) is given by: $C_s = \left[t_{\text{start}} + \Delta,\, t_{\text{end}} -\Delta\right]$, with \(\Delta\) set to 2 seconds. This trimming minimizes phase-transition noise and ensures that each clip delivers a consistent signal for either movement or non-movement (see Figure \ref{sampling2}).

\paragraph{\textbf{Sliding-window sampling}} In the fine-tuning phase, our aim is to enable the model to handle videos of arbitrary length by exposing it to the full spectrum of real-world dynamics. To this end, we employ a sliding window sampling strategy that systematically extracts clips across the entire recording. Unlike the self-supervised phase, these sliding window clips may contain a mix of movement, non-movement, and transitional frames. To accurately capture this variability, we assign probabilistic labels to each clip that reflect the proportion of movement present. For example, a clip containing 70\% movement frames and 30\% non-movement frames is labeled with corresponding probabilities (0.7 for movement and 0.3 for non-movement). Consequently, video recordings are categorized into three distinct regimes: non-movement (label: 0), movement (label: 1), and transitions (represented by continuous values between 0 and 1).

Figure \ref{sampling2} summarizes these two sampling strategies. Performance comparisons between clean-cut and sliding window sampling are detailed in the ablation studies and visualized in Figure \ref{tsne}.

\subsubsection{Data Augmentation}
In our CURL framework, augmentations are carefully designed to preserve essential motion cues while simulating real-world disturbances—such as brief probe movements or fluctuating imaging conditions—to ensure that the learned representations focus on the underlying anatomical and dynamic features.
\paragraph{\textbf{Spatial Augmentation}}
Conventional image-based spatial augmentations—such as brightness and contrast adjustments, rotations, scaling, and noise injection—are typically applied on a frame-by-frame basis. However, independent augmentations across frames can disrupt the temporal coherence that is crucial for accurately capturing motion patterns. To overcome this, we generate the augmentation hyperparameters once per video clip and apply them uniformly across all frames. This approach preserves the spatial integrity of key fetal features (e.g., limb contours and head structures) while maintaining consistent motion cues throughout the clip. Visual examples for a few of these spatial augmentations are provided in Figure \ref{aug_imgs}.

\paragraph{\textbf{Temporal Augmentation}}
Temporal augmentations are critical for enabling the model to learn robust representations that capture the dynamic evolution of fetal movements. Clinical ultrasound videos often exhibit transient disturbances—such as brief probe movements or momentary signal dropouts—which can obscure the true motion patterns. To simulate these conditions without disrupting the intrinsic temporal structure, we employ spatio-temporal augmentations, including techniques such as tube masking and frame masking. By applying fixed random masks consistently across frames \cite{tong2022videomae,feichtenhofer2022masked}, we preserve the overall temporal continuity while compelling the model to focus on dynamic changes rather than redundant, static information. This strategy effectively enhances the model's ability to capture long-range spatio-temporal dependencies and improves its robustness to real-world noise.

Overall, our comprehensive augmentation strategy, combining temporally consistent spatial modifications with informed temporal augmentations—ensures that the CURL framework effectively captures both the anatomical details and dynamic aspects of fetal movements, thereby enhancing model generalization. The impact of these augmentations is validated by our experimental results (see Tables \ref{tab:aug_compare} and \ref{tab:augmentation} and Figure \ref{map}).

\begin{figure*}[]
	\centering
	\includegraphics[width=\textwidth]{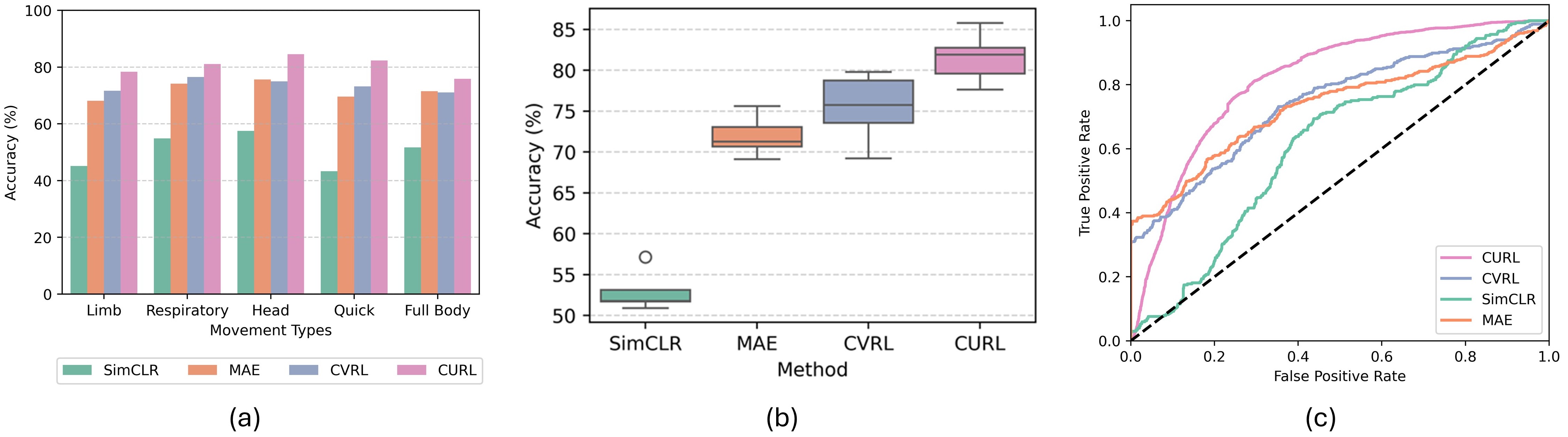}
	\caption{(a) Bar graph comparing the performance on fetal movement subclasses among state-of-the-art representation learning algorithms, (b) Box plots presenting the performance distributions of various representation learning algorithms evaluated using 5-fold cross-validation on the fetal movement dataset and (c) ROC curve comparing the proposed algorithm with other state-of-the-art representation learning methods on our in-house dataset, demonstrating superior classification performance. The model used is ViT-L.}
	\label{box_bar_roc}
\end{figure*}

\subsubsection{Finetuning Protocol}
After the self-supervised training phase, we freeze the pre-trained video encoder and append a lightweight linear classification head to adapt the model for the fetal movement classification task. This linear probing strategy leverages the robust spatio-temporal features previously learned while significantly reducing the number of trainable parameters.

During fine-tuning, a sliding window sampling strategy is employed to extract clips from the full-length ultrasound recordings, ensuring that the entire video is covered regardless of its duration. In contrast to the self-supervised phase—where clips are curated to include only clear movement or non-movement segments—these sliding window samples naturally encompass mixed content, including clear movement, non-movement, and transitional frames. To handle this variability, each clip is assigned probabilistic labels that reflect the proportion of movement versus non-movement frames. For example, a clip composed of 70\% movement frames and 30\% non-movement frames is annotated with corresponding probabilistic outputs as shown in Figure \ref{sampling2}. This approach is analogous to the regularization effect achieved by the MixUp strategy \cite{zhang2017mixup}, where samples from different classes are linearly combined to promote smoother decision boundaries.

The final classification head is optimized using the cross-entropy loss, defined as:

\begin{equation}
L_{CE} = - \frac{1}{N} \sum_{i=1}^{N} \sum_{c=1}^{C} y_{i,c} \log \hat{y}_{i,c}
\end{equation}

where \(N\) denotes the number of training samples, \(C\) is the number of classes (e.g., movement and non-movement), \(y_{i,c}\) represents the ground truth probabilistic label for the \(i\)-th sample, and \(\hat{y}_{i,c}\) is the predicted probability for class \(c\).

This fine-tuning stage, involving the freezing of the pre-trained encoder and the training of only the linear classification layer, is referred to as linear evaluation or linear probing. 

In combination, our informed sampling strategy, targeted augmentations, and efficient linear evaluation form a coherent downstream adaptation pipeline that maximizes the utility of self-supervised pre-training while maintaining computational efficiency.

\section{Experiments and Results}
We evaluated the proposed CURL framework by benchmarking it against leading supervised and self-supervised models for fetal movement detection. First, we summarize the implementation details and validation protocol; then we present the compared methods and quantitative metrics used to assess performance.

\subsection{Experimental Setup and Implementation Details}
For the self-supervised representation learning phase, we employed the AdamW optimizer with a cosine decay learning schedule, starting at a learning rate of 0.003 with a 3-epoch warm-up. A batch size of 128 was used, and gradient accumulation in PyTorch mitigated single-GPU memory constraints.

During the fine-tuning phase, we adopted a linear evaluation protocol: the pre-trained video encoder was frozen, and only a lightweight classification head was trained using SGD (learning rate 0.01, batch size 16). Performance was assessed via patient-wise 5-fold cross-validation, in which all clips from each subject were confined to a single fold to prevent data leakage and simulate realistic generalization. Key metrics included specificity, sensitivity, weighted precision, weighted F1-score, balanced accuracy (bACC), and AUROC. All experiments were conducted on an NVIDIA A6000 GPU with 48 GB of memory.

\subsection{Benchmark Methods}
To contextualize our results, we compare the performance of CURL with several state-of-the-art models that have demonstrated success in human action recognition and fetal movement detection. In the realm of supervised learning, models such as CNN+LSTM  \cite{carreira2017quo} combine convolutional feature extraction with LSTM networks to capture temporal dependencies, while architectures like I3D \cite{carreira2017quo} utilize 3D convolutions for spatiotemporal feature extraction; its Two-Stream variant further integrates separate pathways to independently process spatial and motion cues. Additional supervised models, such as SlowFast and X3D \cite{feichtenhofer2019slowfast,feichtenhofer2020x3d}, employ dual-pathway and computationally efficient 3D convolutional strategies respectively, enabling them to capture both high-resolution spatial details and rapid motion dynamics.

\begin{figure*}[]
	\centering
	\includegraphics[width=1.0\textwidth]{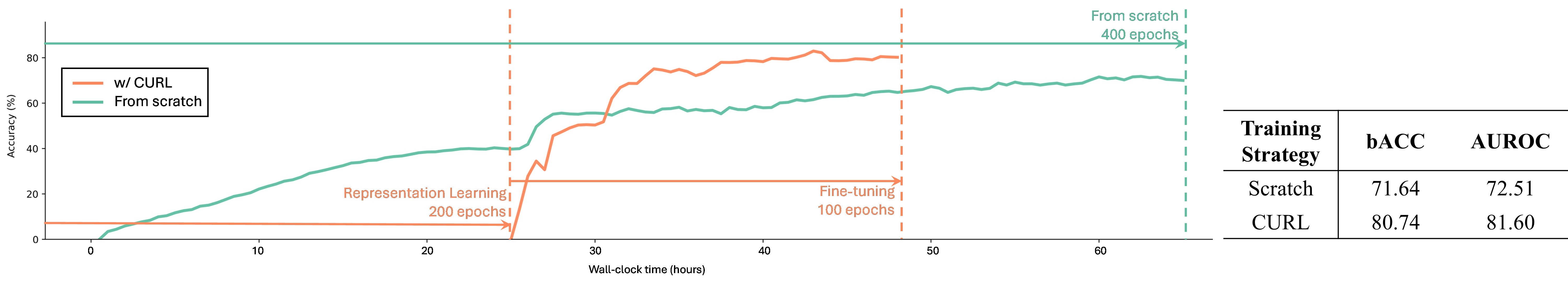}
	\caption{Representation learning with fine-tuning (CURL) outperforms training from scratch in both accuracy and speed. The left panel shows balanced accuracy (bACC) on our in-house dataset over wall-clock training time (measured on a single A6000 GPU), while the right table compares the final bACC and AUROC metrics for models trained from scratch versus those using the proposed CURL algorithm.}
	\label{ab2}
\end{figure*}

Transformer-based architectures have also shown strong performance in video analysis. For example, MViTv2-L \cite{li2022mvitv2} extends the Multiscale Vision Transformer framework by incorporating decomposed relative positional embeddings and residual pooling, whereas VideoSwin-L \cite{liu2022video} adapts the Swin Transformer to video tasks by emphasizing spatial locality and leveraging pre-trained image models. Pure transformer-based models like ViViT-L \cite{arnab2021vivit} process video sequences by extracting spatiotemporal tokens and applying factorized attention mechanisms to handle long-range dependencies effectively.

Self-supervised methods such as SimCLR \cite{chen2020simple} use contrastive learning to maximize similarity between different augmented views of the same image, thereby learning high-quality representations. Extensions of this idea, such as CVRL \cite{qian2021spatiotemporal}, integrate transformer-based architectures to capture spatiotemporal information in videos more effectively, and approaches like MAE (Masked Autoencoders) \cite{feichtenhofer2022masked} apply high-ratio tube masking to promote robust video representation learning. In addition to these general models, our evaluation includes recent methods specifically developed for fetal movement recognition, such as FetalMovNet \cite{turkan2025fetalmovnet}, to highlight the performance of CURL within the context of both general action recognition and domain-specific fetal monitoring.

\begin{table}[]
\centering
\caption{Evaluation of the effectiveness of individual spatial and temporal augmentations applied to a single branch for fetal movement classification.}
\label{tab:augmentation}
\begin{tabular}{@{}l@{\hskip 0.3cm}c@{\hskip 0.3cm}c@{\hskip 0.3cm}c@{}}
\hline
\textbf{Augmentation Strategy} & \textbf{bACC} & \textbf{AUROC} & \textbf{Rank} \\ \hline
\multicolumn{4}{l}{\textbf{Spatial Augmentation}} \\ \hline
Rotate ($\theta \in [-30\degree,30\degree]$)        & 71.53 & 73.67 & 1 \\
Gaussian Noise ($\mu=0,\sigma=0.1$)                  & 69.75 & 71.84 & 3 \\
Contrast ($\times 0.5$--$1.5$)                        & 68.56 & 70.62 & 6 \\
Brightness ($\times 0.5$--$1.5$)                      & 69.15 & 71.23 & 5 \\
Median Blur ($r=3$)                                   & 68.05 & 70.09 & 7 \\ \hline
\multicolumn{4}{l}{\textbf{Temporal Augmentation}} \\ \hline
Tube Masking ($\rho=0\%-30\%$)                        & 69.66 & 71.75 & 4 \\
Frame Masking ($\rho=0\%-30\%$)                        & 67.12 & 69.13 & 8 \\
Random Masking ($\rho=0\%-30\%$)                       & 70.68 & 72.80 & 2 \\ \hline
\end{tabular}
\end{table}

\subsection{Fetal Movement Detection Performance}

A comprehensive evaluation of fetal movement detection performance is presented in Table~\ref{tab:benchmark}, where our proposed model is benchmarked against state-of-the-art approaches. Among fully supervised methods, transformer-based architectures such as ViViT-L achieved the highest AUROC of 74.07\%, highlighting their strength in capturing spatiotemporal dependencies. In contrast, CNN-based models like CNN+LSTM and I3D demonstrated lower sensitivity and specificity, likely due to their limited capacity to model long-range temporal dependencies. Although Two-Stream I3D and SlowFast networks showed moderate improvements by incorporating motion cues, their reliance on pre-trained action recognition models restricted their generalization ability for fetal ultrasound data. Similarly, FetalMovNet’s relatively lower performance can be attributed to its inherent framework limitations, as it operates only on a fixed sequence of 10 frames, employs a simple convolutional architecture with a separate attention module, and relies on a late fusion strategy that may not fully capture the complex dynamics of fetal movements.

In contrast, self-supervised learning methods markedly improved detection performance. Our proposed CURL framework achieved an AUROC of 81.60\%, underscoring the benefits of learning rich video representations in a self-supervised manner. After pre-training, the model was fine-tuned on the labeled data using a linear evaluation protocol, ensuring that the performance gains are directly attributable to the quality of the learned representations. Figure~\ref{box_bar_roc}(a) illustrates that the proposed model maintains consistent performance across different fetal movement subclasses, in contrast to traditional methods which tend to exhibit fluctuations across movement types. This robustness is further evidenced by the box plots in Figure~\ref{box_bar_roc}(b), which show that CURL not only achieves a superior median accuracy but also demonstrates lower standard-deviation across 5-fold cross-validation splits when compared to SimCLR, MAE, and CVRL.

Moreover, the ROC curves depicted in Figure~\ref{box_bar_roc}(c) reveal a significant margin of improvement for CURL over existing self-supervised models, with a higher true positive rate across varying decision thresholds. This performance, combined with its enhanced sensitivity and specificity, confirms that CURL effectively captures both spatial and temporal information, making it a technically robust and clinically valuable solution for ultrasound-based fetal movement monitoring.

\section{Ablation Studies and Discussion}

\subsubsection{Representation Learning vs. Supervised Learning}
To evaluate the advantages of self-supervised representation learning over traditional supervised learning from scratch, we compare models trained using the proposed CURL algorithm with those trained without any pre-training. Figure \ref{ab2} provides a comparison of bACC and AUROC for both training strategies. The results clearly demonstrate that CURL significantly outperforms training from scratch, achieving an absolute improvement of 9.1\% in bACC and 9.09\% in AUROC. These findings highlight the effectiveness of self-supervised pre-training in capturing robust feature representations that lead to superior classification performance.

Additionally, learning curves in Figure \ref{ab2} summarize the comparison between CURL-based training and training from scratch. The x-axis represents the wall-clock training time on a single A6000 GPU, while the y-axis denotes ACC. It is evident that CURL not only achieves higher final accuracy but also converges significantly faster compared to training from scratch. This indicates that leveraging self-supervised representations allows for more efficient training, reducing computational costs while maintaining superior performance. These results confirm the advantages of representation learning over fully supervised learning, particularly when labeled data is limited. By leveraging self-supervised pre-training, CURL provides a strong initialization, allowing for improved generalization and enhanced fetal movement classification performance.

\begin{figure}[]
	\centering
	\includegraphics[width=3.5in]{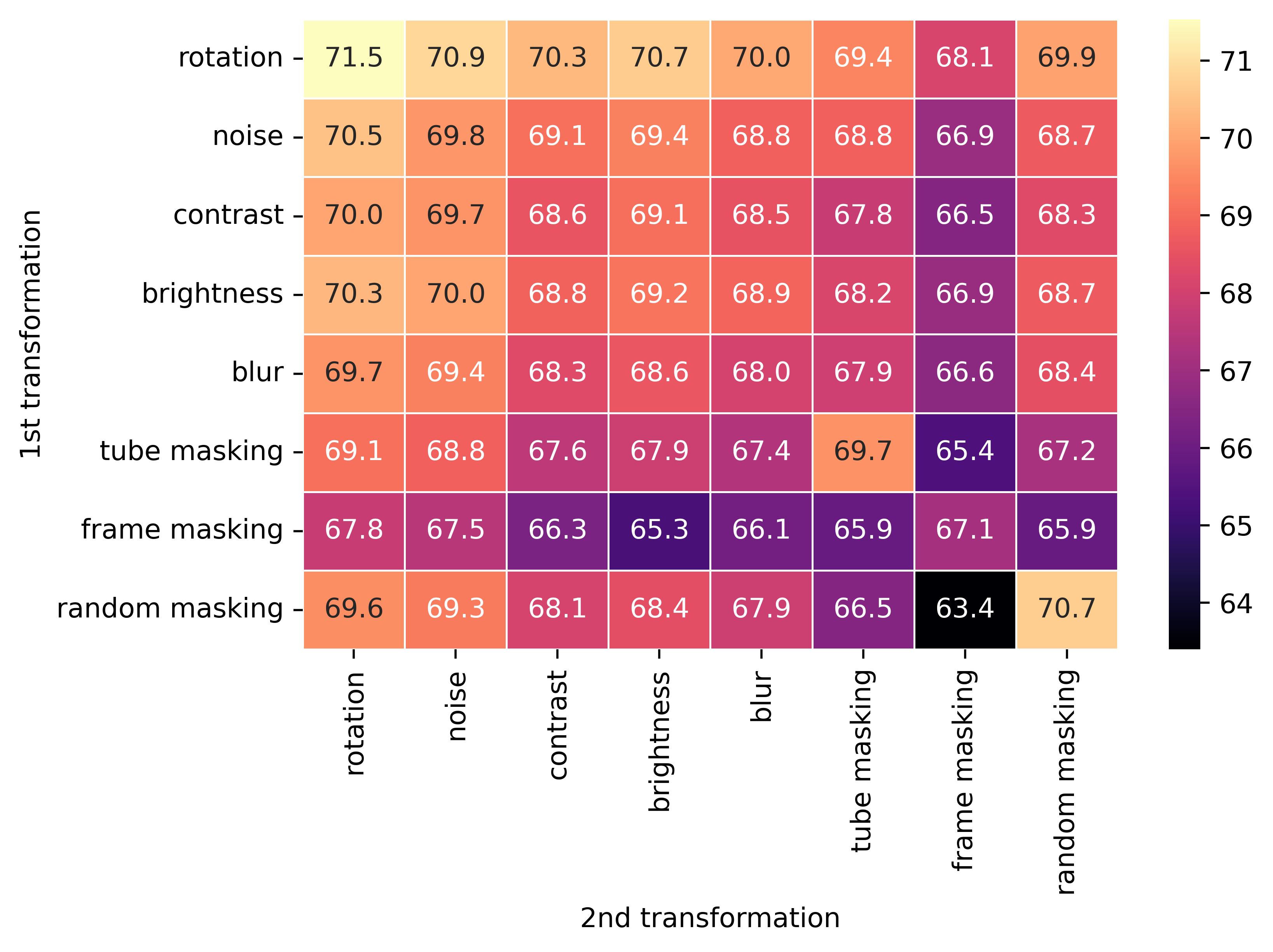}
	\caption{Linear evaluation (in-house dataset accuracy) showing the impact of individual and paired data augmentations, applied only to one branch. Diagonal cells report performance using a single augmentation, while off-diagonal cells display the accuracy (bACC) when two augmentations are sequentially combined.}
	\label{map}
\end{figure}

\subsubsection{Impact of Spatio-Temporal Augmentations}
Data augmentation plays a crucial role in self-supervised learning by enhancing the model's ability to extract robust and invariant features. In this study, we evaluate the effectiveness of both spatial and temporal augmentation strategies for fetal movement classification. Table \ref{tab:augmentation} presents the impact of individual augmentation techniques on model performance, assessed using bACC and AUROC.

Among the spatial augmentations, rotation proves to be the most effective, achieving the highest bACC (71.53\%) and AUROC (73.67\%), indicating that orientation-invariant features are beneficial for fetal movement detection. Gaussian noise and random masking also contribute positively to performance, ranking 2nd and 3rd, respectively. Other spatial augmentations, including contrast, brightness adjustment, and median blur, exhibit comparatively lower improvements.

For temporal augmentations, tube masking, frame masking, and random masking are evaluated. Among these, random masking provides the best results, achieving bACC of 70.68\% and AUROC of 72.80\%, followed closely by tube masking. This suggests that occluding parts of the video sequences in a structured manner aids in learning temporal dynamics while still preserving critical motion cues.

\begin{table}[]
\centering
\caption{Comparison of model performance when using only spatial augmentations, only temporal augmentations, and a combined spatio-temporal approach.}
\label{tab:aug_compare}
\begin{tabular}{cc|c|c}
\hline
\textbf{\begin{tabular}[c]{@{}c@{}}Spatial  \\ Augmentation\end{tabular}} & \textbf{\begin{tabular}[c]{@{}c@{}}Temporal\\ Augmentation\end{tabular}} & \textbf{bACC} & \textbf{AUROC} \\ \hline
\checkmark & - & 77.67 & 78.32 \\
- & \checkmark & 74.78 & 76.62 \\
\checkmark & \checkmark & 80.74 & 81.6 \\ \hline
\end{tabular}
\end{table}

To further investigate the combined effect of spatio-temporal augmentations, we compare three different settings in Table \ref{tab:aug_compare}: (i) only spatial augmentations, (ii) only temporal augmentations, and (iii) a combination of both. The results indicate that spatial augmentations alone achieve better performance than temporal augmentations alone. However, the best performance is obtained when both spatial and temporal augmentations are applied together, yielding bACC of 80.74\% and AUROC of 81.6\%. This highlights the complementary nature of spatial and temporal transformations, reinforcing the importance of augmenting both spatial structure and motion information for enhanced fetal movement classification.

Figure~\ref{map} provides additional insights into the interplay of various augmentation techniques. The main diagonal of the matrix illustrates the baseline performance when a single augmentation is applied to one branch, revealing its standalone impact. In contrast, the off-diagonal elements capture the effect of applying two augmentations in sequence. Generally, the combined performance approximates the average of the individual effects; however, slight deviations arise due to interactions between transformations. Complementary augmentations, such as rotation combined with noise, tend to preserve critical structural information while introducing beneficial variability, whereas overly aggressive or conflicting augmentations may compound errors and degrade performance. This analysis underscores the importance of carefully designing augmentation pipelines to balance trade-offs and synergies, ultimately leading to more robust contrastive learning in the context of fetal movement detection.

\begin{table}[]
\caption{Impact of temporal contrastive loss on model performance, comparing full fine-tuning with linear evaluation.}
\label{tab:lin_eval}
\begin{tabular}{ccc|cc}
\hline
\textbf{Tasks} & \textbf{$L_{sc}$} & \textbf{$L_{tc}$} & \textbf{bACC} & \textbf{AUROC} \\ \hline
\multirow{2}{*}{\begin{tabular}[c]{@{}c@{}}Full\\ Finetuning\end{tabular}} & \checkmark & - & 74.51 & 75.24 \\
 & \checkmark & \checkmark & 78.35 & 79.71 \\ \hline
\multirow{2}{*}{\begin{tabular}[c]{@{}c@{}}Linear\\ Evaluation\end{tabular}} & \checkmark & - & 75.87 & 76.46 \\
 & \checkmark & \checkmark & 80.74 & 81.60 \\ \hline
\end{tabular}
\end{table}

\subsubsection{Dual-Contrastive Loss: Fine-Tuning vs. Linear Evaluation}
To assess the impact of \textit{temporal contrastive loss} on fetal movement classification, we compare two training paradigms, the full fine-tuning and the linear evaluation. Table~\ref{tab:lin_eval} presents the performance comparison for both strategies under different loss configurations.  

In the full fine-tuning setting, the video encoder is updated along with the classification head. When using only the spatial contrastive loss (\(L_{sc}\)), the model achieves bACC of 74.51\% and AUROC of 75.24\%. However, incorporating the temporal contrastive loss (\(L_{tc}\)) significantly boosts performance, reaching bACC of 78.35\% and AUROC of 79.71\%, confirming the importance of leveraging temporal dependencies.  

For the linear evaluation setup, where the video encoder is frozen and only a classification head is trained, we observe a similar trend. The model trained with only \(L_{sc}\) attains a bACC of 75.87\% and AUROC of 76.46\%, surpassing the fine-tuning counterpart. When both \(L_{sc}\) and \(L_{tc}\) are employed, performance further improves to 80.74\% bACC and 81.60\% AUROC, marking the best results across all configurations.   These findings demonstrate that linear evaluation with dual-contrastive loss outperforms full fine-tuning, indicating that the pretrained representations are already well-structured, and extensive fine-tuning is not required for optimal performance. The effectiveness of \(L_{tc}\) further highlights the importance of temporal consistency in learning robust fetal movement features.

\begin{figure}[]
	\centering
	\includegraphics[width=0.5\textwidth]{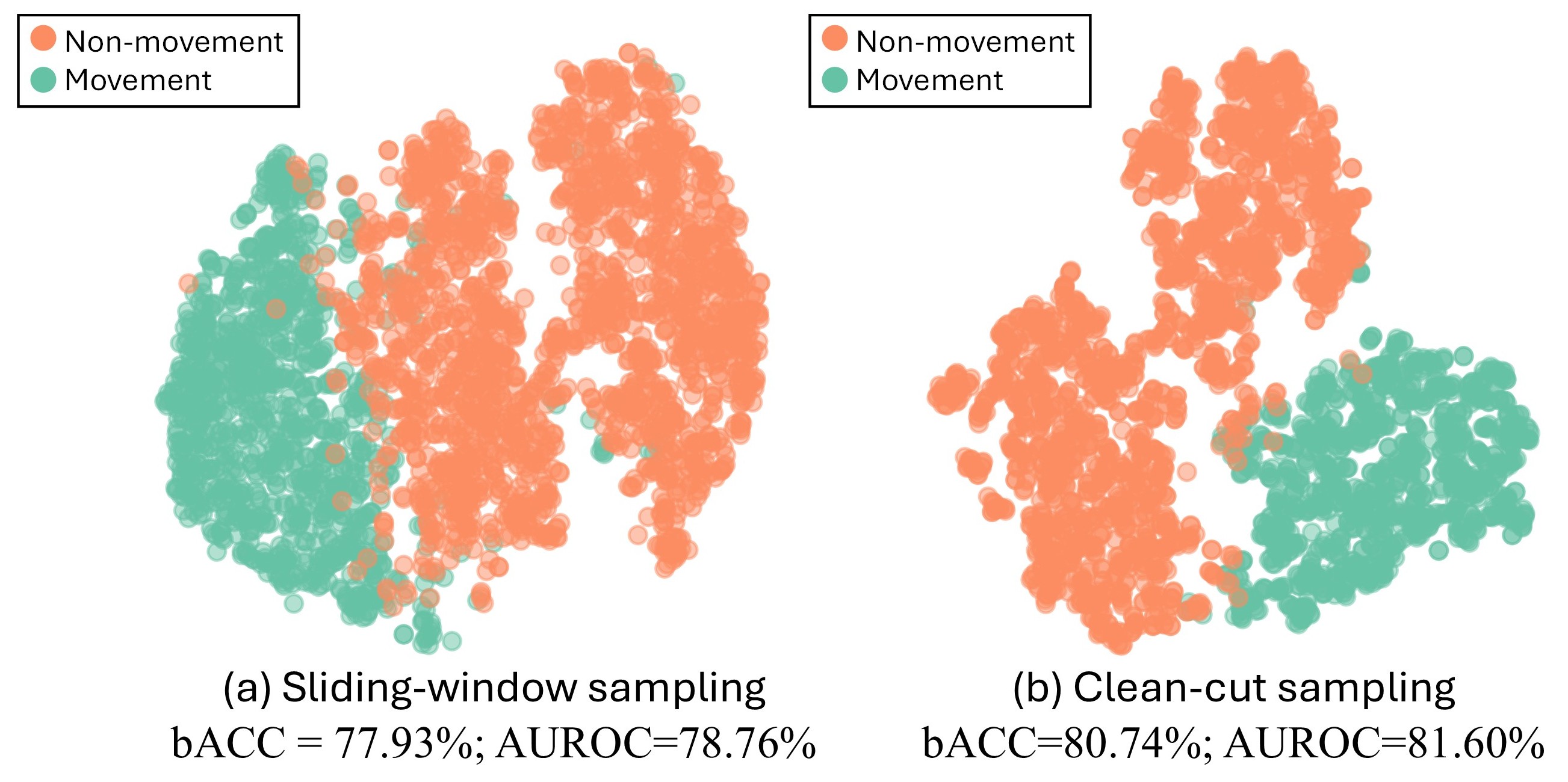}
	\caption{2D t-SNE visualizations of feature representations learned using (a) sliding-window sampling and (b) clean-cut sampling strategies. The clean-cut sampling approach results in more distinct clusters and improved classification accuracy, underscoring its effectiveness in capturing temporal dynamics in fetal movement.}
	\label{tsne}
\end{figure}

\subsection{Evaluation of Sampling Strategies}

To analyze the impact of different temporal sampling strategies on fetal movement classification, we compare sliding-window sampling and clean-cut sampling for representation learning phase. The performance results, in terms of balanced accuracy (bACC) and AUROC, demonstrate the advantage of the clean-cut strategy over the sliding-window approach. The sliding-window sampling yields a bACC of 77.93\% and AUROC of 78.76\%, whereas the clean-cut sampling achieves a significantly higher bACC of 80.74\% and AUROC of 81.60\%.  

Figure~\ref{tsne} presents t-SNE visualizations \cite{van2008visualizing} of feature representations learned using the two sampling strategies. The results clearly show that clean-cut sampling enables the video encoder to learn more robust and generalized features, leading to distinct and well-separated clusters corresponding to different fetal movement classes. This enhanced separability can be largely attributed to the removal of transitional frames during the self-supervised learning phase, which effectively reduces noise and ambiguity in the training data. Overall, the superior performance of the clean-cut sampling strategy underscores its effectiveness in preserving the inherent spatio-temporal structure of fetal movements, ultimately resulting in more reliable feature representations and improved classification accuracy.


\subsection{Discussion and Future Directions}
\begin{figure}[]
	\centering
	\includegraphics[width=0.45\textwidth]{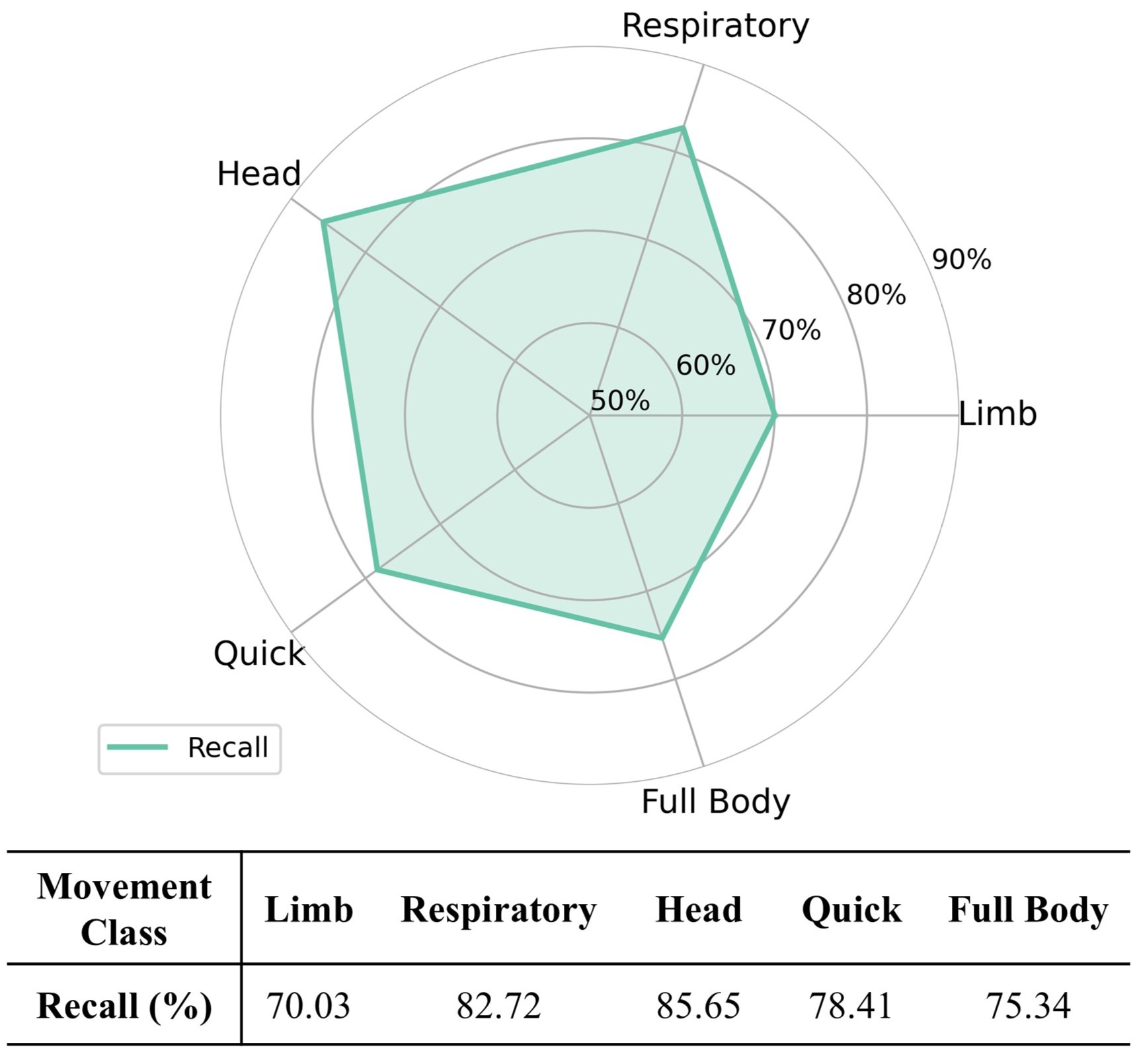}
	\caption{Radar plot showing individual recall rates for each fetal movement class.}
	\label{rad}
\end{figure}

We assessed class-specific recall rates (Figure~\ref{rad}), revealing performance variability across movement types and identifying both model strengths and key limitations. The model achieved the highest recall for head movements (85.65\%), likely due to their distinctive ultrasound characteristics and adequate duration (average of 5.38 seconds), particularly evident in the third trimester. Respiratory movements also exhibited high recall (82.72\%), benefiting from their frequent occurrence and extended duration—panting episodes averaging 35.11 seconds and hiccups approximately 5.48 seconds—yielding robust training examples.

Quick movements, despite brief durations (average 1.35 seconds for startles), achieved moderately high recall (78.41\%), reflecting their distinctive, easily identifiable patterns. Conversely, full-body movements, such as Prechtl (22.76 seconds) and trunk motions (7.56 seconds), showed slightly lower recall (75.34\%) due to occasional confusion with ultrasound artifacts like probe movements and noise. Isolated limb movements demonstrated the lowest recall (70.03\%), attributed to their subtle nature, small anatomical size, and frequent occlusion \cite{zhao2023lower}. This highlights a key limitation: the model's reduced sensitivity to subtle, isolated limb motions that are clinically significant yet technically challenging to detect.

Clinically, these performance differences align with findings by Stanojević et al. \cite{stanojevic2023fetal}, emphasizing that comprehensive assessment of general movements (GMs) using holistic perceptual frameworks ("gestalt" perception) is crucial for predicting neurodevelopmental outcomes. The observed performance disparities underscore the importance of refining limb-specific detection to enhance the model's clinical utility.

Future research should address these limitations through targeted data augmentation strategies focused on subtle limb movements, refinement of feature extraction methods, and the expansion of annotated datasets. Such improvements would further strengthen the model’s clinical applicability and diagnostic reliability.

\section*{Conclusion}
In this study, we introduced the Contrastive Ultrasound Video Representation Learning (CURL) framework for automated fetal movement detection using extended ultrasound recordings. CURL employs a dual contrastive loss strategy, effectively capturing the intricate spatio-temporal dynamics of fetal motion, and addresses limitations inherent in traditional subjective assessment methods. The framework incorporates carefully designed sampling strategies, notably the clean-cut approach, to optimize learning by reducing transitional noise.

Experimental evaluations demonstrate CURL’s robust performance, particularly in detecting clinically relevant movements such as head (85.65\%) and respiratory patterns (82.72\%), alongside competent performance on full-body (75.34\%) and quick movements (78.41\%). However, the model’s lower recall for isolated limb movements (70.03\%) highlights an important limitation, emphasizing the need for targeted improvements.

Overall, CURL represents a promising non-invasive, objective tool for fetal movement analysis, offering substantial potential to enhance prenatal monitoring and early detection of developmental anomalies. Future research directions include refining limb-specific movement detection, augmenting the training dataset with diverse clinical samples, and validating model performance across broader populations, facilitating seamless integration into routine clinical practice.

\section*{Acknowledgment}

W. Cheng, F. Marzbanrad, and L. W. Yap gratefully acknowledge financial support from the National Health and Medical Research Council (NHMRC) through the Ideas Grant scheme (APP2004444). W. Cheng also acknowledges support from the NHMRC Investigator Grant (APP2010154). S. Gong is supported by the Jack Brockhoff Foundation (Grant No. 4659-2019).

\section*{References}
\bibliographystyle{IEEEtran}
\bibliography{main}

\begin{thebibliography}{10}
\providecommand{\url}[1]{#1}
\csname url@samestyle\endcsname
\providecommand{\newblock}{\relax}
\providecommand{\bibinfo}[2]{#2}
\providecommand{\BIBentrySTDinterwordspacing}{\spaceskip=0pt\relax}
\providecommand{\BIBentryALTinterwordstretchfactor}{4}
\providecommand{\BIBentryALTinterwordspacing}{\spaceskip=\fontdimen2\font plus
\BIBentryALTinterwordstretchfactor\fontdimen3\font minus \fontdimen4\font\relax}
\providecommand{\BIBforeignlanguage}[2]{{%
\expandafter\ifx\csname l@#1\endcsname\relax
\typeout{** WARNING: IEEEtran.bst: No hyphenation pattern has been}%
\typeout{** loaded for the language `#1'. Using the pattern for}%
\typeout{** the default language instead.}%
\else
\language=\csname l@#1\endcsname
\fi
#2}}
\providecommand{\BIBdecl}{\relax}
\BIBdecl

\bibitem{1}
N.~Melamed, A.~Baschat, Y.~Yinon, A.~Athanasiadis, F.~Mecacci, F.~Figueras, V.~Berghella, A.~Nazareth, M.~Tahlak, H.~D. McIntyre \emph{et~al.}, ``Figo (international federation of gynecology and obstetrics) initiative on fetal growth: best practice advice for screening, diagnosis, and management of fetal growth restriction,'' \emph{International Journal of Gynaecology and Obstetrics}, vol. 152, no. Suppl 1, p.~3, 2021.

\bibitem{2}
N.~E. Skakkeb{\ae}k, R.~Lindahl-Jacobsen, H.~Levine, A.-M. Andersson, N.~J{\o}rgensen, K.~M. Main, {\O}.~Lidegaard, L.~Priskorn, S.~A. Holmboe, E.~V. Br{\"a}uner \emph{et~al.}, ``Environmental factors in declining human fertility,'' \emph{Nature Reviews Endocrinology}, vol.~18, no.~3, pp. 139--157, 2022.

\bibitem{3}
S.~Naja, M.~Makhlouf, and M.~A.~H. Chehab, ``An ageing world of the 21st century: a literature review,'' \emph{Int J Community Med Public Health}, vol.~4, no.~12, pp. 4363--9, 2017.

\bibitem{4}
J.~Lai, N.~C. Nowlan, R.~Vaidyanathan, C.~J. Shaw, and C.~C. Lees, ``Fetal movements as a predictor of health,'' \emph{Acta obstetricia et gynecologica Scandinavica}, vol.~95, no.~9, pp. 968--975, 2016.

\bibitem{5}
A.~G. Olesen and J.~A. Svare, ``Decreased fetal movements: background, assessment, and clinical management,'' \emph{Acta obstetricia et gynecologica Scandinavica}, vol.~83, no.~9, pp. 818--826, 2004.

\bibitem{dutton2012predictors}
P.~J. Dutton, L.~K. Warrander, S.~A. Roberts, G.~Bernatavicius, L.~M. Byrd, D.~Gaze, J.~Kroll, R.~L. Jones, C.~P. Sibley, J.~F. Fr{\o}en \emph{et~al.}, ``Predictors of poor perinatal outcome following maternal perception of reduced fetal movements--a prospective cohort study,'' \emph{PloS one}, vol.~7, no.~7, p. e39784, 2012.

\bibitem{wu2013research}
S.~Wu, Y.~Shen, Z.~Zhou, L.~Lin, Y.~Zeng, and X.~Gao, ``Research of fetal ecg extraction using wavelet analysis and adaptive filtering,'' \emph{Computers in biology and medicine}, vol.~43, no.~10, pp. 1622--1627, 2013.

\bibitem{abeywardhana2018time}
S.~Abeywardhana, H.~Subhashini, W.~Wasalaarachchi, G.~Wimalarathna, M.~Ekanayake, G.~Godaliyadda, J.~Wijayakulasooriya, and R.~Rathnayake, ``Time domain analysis for fetal movement detection using accelerometer data,'' in \emph{2018 IEEE region 10 humanitarian technology conference (R10-HTC)}.\hskip 1em plus 0.5em minus 0.4em\relax IEEE, 2018, pp. 1--5.

\bibitem{xu2023development}
S.~Xu, ``Development of a portable home fetal heart rate monitor that connects to mobile phones,'' in \emph{Proceedings of the 2023 7th International Conference on Electronic Information Technology and Computer Engineering}, 2023, pp. 1746--1752.

\bibitem{zhu2016understanding}
M.-Y. Zhu, ``Understanding freertos: A requirement analysis,'' \emph{CoreTek Syst., Inc., Beijing, China, Tech. Rep}, 2016.

\bibitem{nishihara2008long}
K.~Nishihara, S.~Horiuchi, H.~Eto, and M.~Honda, ``A long-term monitoring of fetal movement at home using a newly developed sensor: An introduction of maternal micro-arousals evoked by fetal movement during maternal sleep,'' \emph{Early human development}, vol.~84, no.~9, pp. 595--603, 2008.

\bibitem{mesbah2021automatic}
M.~Mesbah, M.~S. Khlif, S.~Layeghy, C.~E. East, S.~Dong, A.~Brodtmann, P.~B. Colditz, and B.~Boashash, ``Automatic fetal movement recognition from multi-channel accelerometry data,'' \emph{Computer methods and programs in biomedicine}, vol. 210, p. 106377, 2021.

\bibitem{avci2017tracking}
R.~Avci, J.~D. Wilson, D.~Escalona-Vargas, and H.~Eswaran, ``Tracking fetal movement through source localization from multisensor magnetocardiographic recordings,'' \emph{IEEE journal of biomedical and health informatics}, vol.~22, no.~3, pp. 758--765, 2017.

\bibitem{lutter2011indices}
W.~J. Lutter and R.~T. Wakai, ``Indices and detectors for fetal mcg actography,'' \emph{IEEE transactions on biomedical engineering}, vol.~58, no.~6, pp. 1874--1880, 2011.

\bibitem{stanojevic2023fetal}
M.~Stanojevi{\'c}, S.~Malinac, A.~Kurjak, and E.~Medjedovi{\'c}, ``From fetal to neonatal neurobehavior,'' \emph{Ultrasound Obstet Gynecol}, vol.~17, no.~4, pp. 323--331, 2023.

\bibitem{lai2018performance}
J.~Lai, R.~Woodward, Y.~Alexandrov, Q.~ain Munnee, C.~C. Lees, R.~Vaidyanathan, and N.~C. Nowlan, ``Performance of a wearable acoustic system for fetal movement discrimination,'' \emph{PloS one}, vol.~13, no.~5, p. e0195728, 2018.

\bibitem{alfirevic2017continuous}
Z.~Alfirevic, G.~M. Gyte, A.~Cuthbert, and D.~Devane, ``Continuous cardiotocography (ctg) as a form of electronic fetal monitoring (efm) for fetal assessment during labour,'' \emph{Cochrane database of systematic reviews}, no.~2, 2017.

\bibitem{ryo2012new}
E.~Ryo, K.~Nishihara, S.~Matsumoto, and H.~Kamata, ``A new method for long-term home monitoring of fetal movement by pregnant women themselves,'' \emph{Medical engineering \& physics}, vol.~34, no.~5, pp. 566--572, 2012.

\bibitem{boashash2014passive}
B.~Boashash, M.~S. Khlif, T.~Ben-Jabeur, C.~E. East, and P.~B. Colditz, ``Passive detection of accelerometer-recorded fetal movements using a time--frequency signal processing approach,'' \emph{Digital Signal Processing}, vol.~25, pp. 134--155, 2014.

\bibitem{delay2021novel}
U.~Delay, T.~Nawarathne, S.~Dissanayake, S.~Gunarathne, T.~Withanage, R.~Godaliyadda, C.~Rathnayake, P.~Ekanayake, and J.~Wijayakulasooriya, ``Novel non-invasive in-house fabricated wearable system with a hybrid algorithm for fetal movement recognition,'' \emph{Plos one}, vol.~16, no.~7, p. e0254560, 2021.

\bibitem{nageotte_2015}
\BIBentryALTinterwordspacing
M.~P. Nageotte, ``Fetal heart rate monitoring,'' \emph{Seminars in Fetal and Neonatal Medicine}, vol.~20, no.~3, p. 144–148, Mar 2015. [Online]. Available: \url{https://www.sciencedirect.com/topics/medicine-and-dentistry/fetus-electrocardiography}
\BIBentrySTDinterwordspacing

\bibitem{marzbanrad2018cardiotocography}
F.~Marzbanrad, L.~Stroux, and G.~D. Clifford, ``Cardiotocography and beyond: a review of one-dimensional doppler ultrasound application in fetal monitoring,'' \emph{Physiological measurement}, vol.~39, no.~8, p. 08TR01, 2018.

\bibitem{rooijakkers2014fetal}
M.~J. Rooijakkers, H.~de~Lau, C.~Rabotti, S.~G. Oei, J.~W. Bergmans, and M.~Mischi, ``Fetal movement detection based on qrs amplitude variations in abdominal ecg recordings,'' in \emph{2014 36th Annual International Conference of the IEEE Engineering in Medicine and Biology Society}.\hskip 1em plus 0.5em minus 0.4em\relax IEEE, 2014, pp. 1452--1455.

\bibitem{liu2014multi}
C.~Liu, P.~Li, C.~Di~Maria, L.~Zhao, H.~Zhang, and Z.~Chen, ``A multi-step method with signal quality assessment and fine-tuning procedure to locate maternal and fetal qrs complexes from abdominal ecg recordings,'' \emph{Physiological measurement}, vol.~35, no.~8, p. 1665, 2014.

\bibitem{rooijakkers2015feasibility}
M.~J. Rooijakkers, C.~Rabotti, H.~de~Lau, S.~G. Oei, J.~W. Bergmans, and M.~Mischi, ``Feasibility study of a new method for low-complexity fetal movement detection from abdominal ecg recordings,'' \emph{IEEE Journal of Biomedical and Health Informatics}, vol.~20, no.~5, pp. 1361--1368, 2015.

\bibitem{shokouhmand2022fetal}
A.~Shokouhmand and N.~Tavassolian, ``Fetal electrocardiogram extraction using dual-path source separation of single-channel non-invasive abdominal recordings,'' \emph{IEEE Transactions on Biomedical Engineering}, vol.~70, no.~1, pp. 283--295, 2022.

\bibitem{govindan2011novel}
R.~Govindan, S.~Vairavan, U.~Ulusar, J.~Wilson, S.~McKelvey, H.~Preissl, and H.~Eswaran, ``A novel approach to track fetal movement using multi-sensor magnetocardiographic recordings,'' \emph{Annals of biomedical engineering}, vol.~39, pp. 964--972, 2011.

\bibitem{stinstra2002influence}
J.~Stinstra and M.~Peters, ``The influence of fetoabdominal tissues on fetal ecgs and mcgs,'' \emph{Archives of physiology and biochemistry}, vol. 110, no.~3, pp. 165--176, 2002.

\bibitem{yaqub2012automatic}
M.~Yaqub, R.~Napolitano, C.~Ioannou, A.~T. Papageorghiou, and J.~A. Noble, ``Automatic detection of local fetal brain structures in ultrasound images,'' in \emph{2012 9th IEEE International Symposium on Biomedical Imaging (ISBI)}.\hskip 1em plus 0.5em minus 0.4em\relax IEEE, 2012, pp. 1555--1558.

\bibitem{sobhaninia2019fetal}
Z.~Sobhaninia, S.~Rafiei, A.~Emami, N.~Karimi, K.~Najarian, S.~Samavi, and S.~R. Soroushmehr, ``Fetal ultrasound image segmentation for measuring biometric parameters using multi-task deep learning,'' in \emph{2019 41st annual international conference of the IEEE engineering in medicine and biology society (EMBC)}.\hskip 1em plus 0.5em minus 0.4em\relax IEEE, 2019, pp. 6545--6548.

\bibitem{lei2014automatic}
B.~Lei, L.~Zhuo, S.~Chen, S.~Li, D.~Ni, and T.~Wang, ``Automatic recognition of fetal standard plane in ultrasound image,'' in \emph{2014 IEEE 11th International Symposium on Biomedical Imaging (ISBI)}.\hskip 1em plus 0.5em minus 0.4em\relax IEEE, 2014, pp. 85--88.

\bibitem{yu2016fetal}
Z.~Yu, D.~Ni, S.~Chen, S.~Li, T.~Wang, and B.~Lei, ``Fetal facial standard plane recognition via very deep convolutional networks,'' in \emph{2016 38th annual international conference of the IEEE Engineering in Medicine and Biology Society (EMBC)}.\hskip 1em plus 0.5em minus 0.4em\relax IEEE, 2016, pp. 627--630.

\bibitem{ishikawa2019detecting}
G.~Ishikawa, R.~Xu, J.~Ohya, and H.~Iwata, ``Detecting a fetus in ultrasound images using grad cam and locating the fetus in the uterus.'' in \emph{ICPRAM}, 2019, pp. 181--189.

\bibitem{van2019automated}
T.~L. van~den Heuvel, H.~Petros, S.~Santini, C.~L. de~Korte, and B.~van Ginneken, ``Automated fetal head detection and circumference estimation from free-hand ultrasound sweeps using deep learning in resource-limited countries,'' \emph{Ultrasound in medicine \& biology}, vol.~45, no.~3, pp. 773--785, 2019.

\bibitem{dozen2020image}
A.~Dozen, M.~Komatsu, A.~Sakai, R.~Komatsu, K.~Shozu, H.~Machino, S.~Yasutomi, T.~Arakaki, K.~Asada, S.~Kaneko \emph{et~al.}, ``Image segmentation of the ventricular septum in fetal cardiac ultrasound videos based on deep learning using time-series information,'' \emph{Biomolecules}, vol.~10, no.~11, p. 1526, 2020.

\bibitem{ravishankar2016hybrid}
H.~Ravishankar, S.~M. Prabhu, V.~Vaidya, and N.~Singhal, ``Hybrid approach for automatic segmentation of fetal abdomen from ultrasound images using deep learning,'' in \emph{2016 IEEE 13th international symposium on biomedical imaging (ISBI)}.\hskip 1em plus 0.5em minus 0.4em\relax IEEE, 2016, pp. 779--782.

\bibitem{chen2020automatic}
X.~Chen, M.~He, T.~Dan, N.~Wang, M.~Lin, L.~Zhang, J.~Xian, H.~Cai, and H.~Xie, ``Automatic measurements of fetal lateral ventricles in 2d ultrasound images using deep learning,'' \emph{Frontiers in neurology}, vol.~11, p. 526, 2020.

\bibitem{arnaout2021ensemble}
R.~Arnaout, L.~Curran, Y.~Zhao, J.~C. Levine, E.~Chinn, and A.~J. Moon-Grady, ``An ensemble of neural networks provides expert-level prenatal detection of complex congenital heart disease,'' \emph{Nature medicine}, vol.~27, no.~5, pp. 882--891, 2021.

\bibitem{zhao2023lower}
X.~Zhao, J.~Awrejcewicz, J.~Li, Y.~He, and Y.~Gu, ``The lower limb movements of the fetus in uterus: A narrative review,'' \emph{Applied Bionics and Biomechanics}, vol. 2023, no.~1, p. 4324889, 2023.

\bibitem{inubashiri2021new}
E.~Inubashiri, S.~Fujita, S.~Shimakura, M.~Kurasawa, T.~Yamamoto, Y.~Watanabe, K.~Deguchi, N.~Akutagawa, K.~Kuroki, and N.~Maeda, ``A new approach for quantitative assessment of fetal general movements in the early second trimester of pregnancy using four-dimensional ultrasound,'' \emph{Journal of Medical Ultrasonics}, vol.~48, no.~3, pp. 335--344, 2021.

\bibitem{dandil2021fetal}
E.~Dand{\i}l, M.~Turkan, F.~E. Urfal{\i}, {\.I}.~Biyik, and M.~Korkmaz, ``Fetal movement detection and anatomical plane recognition using yolov5 network in ultrasound scans,'' \emph{Avrupa Bilim ve Teknoloji Dergisi}, no.~26, pp. 208--216, 2021.

\bibitem{turkan2025fetalmovnet}
M.~Turkan, E.~Dandil, F.~E. Urfali, and M.~Korkmaz, ``Fetalmovnet: A novel deep learning model based on attention mechanism for fetal movement classification in us,'' \emph{IEEE Access}, 2025.

\bibitem{turkan2023automatic}
M.~Turkan, F.~E. Urfal{\i}, and E.~Dand{\i}l, ``Automatic fetal motion detection from trajectory of us videos based on yolov5 and lstm,'' in \emph{Explainable Machine Learning for Multimedia Based Healthcare Applications}.\hskip 1em plus 0.5em minus 0.4em\relax Springer, 2023, pp. 1--20.

\bibitem{heMaskedAutoencodersAre2021}
K.~He, X.~Chen, S.~Xie, Y.~Li, P.~Doll{\'a}r, and R.~Girshick, ``Masked {{Autoencoders Are Scalable Vision Learners}},'' Dec. 2021.

\bibitem{feichtenhoferMaskedAutoencodersSpatiotemporal2022}
C.~Feichtenhofer, H.~Fan, Y.~Li, and K.~He, ``Masked {{Autoencoders As Spatiotemporal Learners}},'' Oct. 2022.

\bibitem{sun2025masked}
F.~Sun and W.~Jin, ``A masked autoencoder network for spatiotemporal predictive learning,'' \emph{Applied Intelligence}, vol.~55, no.~5, p. 348, 2025.

\bibitem{chen2020simple}
T.~Chen, S.~Kornblith, M.~Norouzi, and G.~Hinton, ``A simple framework for contrastive learning of visual representations,'' in \emph{International conference on machine learning}.\hskip 1em plus 0.5em minus 0.4em\relax PmLR, 2020, pp. 1597--1607.

\bibitem{oord2018representation}
A.~v.~d. Oord, Y.~Li, and O.~Vinyals, ``Representation learning with contrastive predictive coding,'' \emph{arXiv preprint arXiv:1807.03748}, 2018.

\bibitem{zhang2021supporting}
D.~Zhang, F.~Nan, X.~Wei, S.~Li, H.~Zhu, K.~McKeown, R.~Nallapati, A.~Arnold, and B.~Xiang, ``Supporting clustering with contrastive learning,'' \emph{arXiv preprint arXiv:2103.12953}, 2021.

\bibitem{chakraborty2020entropy}
S.~Chakraborty, D.~Paul, S.~Das, and J.~Xu, ``Entropy regularized power k-means clustering,'' \emph{arXiv preprint arXiv:2001.03452}, 2020.

\bibitem{carreira2017quo}
J.~Carreira and A.~Zisserman, ``Quo vadis, action recognition? a new model and the kinetics dataset,'' in \emph{proceedings of the IEEE Conference on Computer Vision and Pattern Recognition}, 2017, pp. 6299--6308.

\bibitem{feichtenhofer2019slowfast}
C.~Feichtenhofer, H.~Fan, J.~Malik, and K.~He, ``Slowfast networks for video recognition,'' in \emph{Proceedings of the IEEE/CVF international conference on computer vision}, 2019, pp. 6202--6211.

\bibitem{feichtenhofer2020x3d}
C.~Feichtenhofer, ``X3d: Expanding architectures for efficient video recognition,'' in \emph{Proceedings of the IEEE/CVF conference on computer vision and pattern recognition}, 2020, pp. 203--213.

\bibitem{li2022mvitv2}
Y.~Li, C.-Y. Wu, H.~Fan, K.~Mangalam, B.~Xiong, J.~Malik, and C.~Feichtenhofer, ``Mvitv2: Improved multiscale vision transformers for classification and detection,'' in \emph{Proceedings of the IEEE/CVF conference on computer vision and pattern recognition}, 2022, pp. 4804--4814.

\bibitem{liu2022video}
Z.~Liu, J.~Ning, Y.~Cao, Y.~Wei, Z.~Zhang, S.~Lin, and H.~Hu, ``Video swin transformer,'' in \emph{Proceedings of the IEEE/CVF conference on computer vision and pattern recognition}, 2022, pp. 3202--3211.

\bibitem{arnab2021vivit}
A.~Arnab, M.~Dehghani, G.~Heigold, C.~Sun, M.~Lu{\v{c}}i{\'c}, and C.~Schmid, ``Vivit: A video vision transformer,'' in \emph{Proceedings of the IEEE/CVF international conference on computer vision}, 2021, pp. 6836--6846.

\bibitem{feichtenhofer2022masked}
C.~Feichtenhofer, Y.~Li, K.~He \emph{et~al.}, ``Masked autoencoders as spatiotemporal learners,'' \emph{Advances in neural information processing systems}, vol.~35, pp. 35\,946--35\,958, 2022.

\bibitem{qian2021spatiotemporal}
R.~Qian, T.~Meng, B.~Gong, M.-H. Yang, H.~Wang, S.~Belongie, and Y.~Cui, ``Spatiotemporal contrastive video representation learning,'' in \emph{Proceedings of the IEEE/CVF conference on computer vision and pattern recognition}, 2021, pp. 6964--6974.

\bibitem{tong2022videomae}
Z.~Tong, Y.~Song, J.~Wang, and L.~Wang, ``Videomae: Masked autoencoders are data-efficient learners for self-supervised video pre-training,'' \emph{Advances in neural information processing systems}, vol.~35, pp. 10\,078--10\,093, 2022.

\bibitem{zhang2017mixup}
H.~Zhang, M.~Cisse, Y.~N. Dauphin, and D.~Lopez-Paz, ``mixup: Beyond empirical risk minimization,'' \emph{arXiv preprint arXiv:1710.09412}, 2017.

\bibitem{van2008visualizing}
L.~Van~der Maaten and G.~Hinton, ``Visualizing data using t-sne.'' \emph{Journal of machine learning research}, vol.~9, no.~11, 2008.

\end{thebibliography}

\end{document}